\documentclass[letterpaper,journal]{IEEEtran}
\newif\ifthesis
\usepackage{graphicx} 
\usepackage{amsmath}
\usepackage{amsfonts}
\usepackage{enumitem}
\usepackage{pdflscape}
\usepackage{hyperref}
\usepackage{soul}
\usepackage{color}
\usepackage{subcaption}
\usepackage[skip=0pt]{caption}
\usepackage{bm}
\hypersetup{
  pdftitle={Complete Autonomous Robotic Nasopharyngeal Swab System with Evaluation on a Stochastically Moving Phantom Head},
                              pdfauthor={Peter Q. Lee, John S. Zelek, and Katja Mombaur},                             colorlinks=true,           linkcolor=red,              citecolor=green,            filecolor=magenta,          urlcolor=cyan,               }

\title{Complete Autonomous Robotic Nasopharyngeal Swab System with Evaluation on a Stochastically Moving Phantom Head}
\author{Peter Q. Lee$^{1,*}$, John S. Zelek$^1$ and Katja Mombaur$^{1,2,3}$
\thanks{1. Systems Design Engineering, University of Waterloo, 200 University Avenue West, Waterloo, Canada. 2. Mechanical and Mechatronics Engineering, University of Waterloo, 200 University Avenue
  West, Waterloo, Canada. 3. Optimization and Biomechanics for Human-Centred Robotics (BioRobotics Lab), Institute for Anthropomatics and Robotics, Karlsruhe Institute of Technology, Karlsruhe, Germany *pqjlee@uwaterloo.ca The research presented in this manuscript was first presented in chapter 2 \& 7 of PL's PhD thesis.}}

\begin{document}

   \maketitle
   \begin{abstract}
     The application of autonomous robotics to close-contact healthcare tasks has a clear role for the future due to its potential to reduce infection risks to staff and improve clinical efficiency.
                    Nasopharyngeal (NP) swab sample collection for diagnosing upper-respiratory illnesses is one type of close contact task that is interesting for robotics due to the dexterity requirements and the unobservability of the nasal cavity.
     We propose a control system that performs the test using a collaborative manipulator arm with an instrumented end-effector to take visual and force measurements, under the scenario that the patient is unrestrained and the tools are general enough to be applied to other close contact tasks.
     The system employs a visual servo controller to align the swab with the nostrils. A compliant joint velocity controller inserts the swab along a trajectory optimized through a simulation environment, that also reacts to measured forces applied to the swab.
          Additional subsystems include a fuzzy logic system for detecting when the swab reaches the nasopharynx and a method for detaching the swab and aborting the procedure if safety criteria is violated.
     The system is evaluated using a second robotic arm that holds a nasal cavity phantom and simulates the natural head motions that could occur during the procedure.
     Through extensive experiments, we identify controller configurations capable of effectively performing the NP swab test even with significant head motion, which demonstrates the safety and reliability of the system.
         \end{abstract}

\section{Introduction}
\label{sec:int:intro}
\begin{figure*}
  \includegraphics[width=\linewidth]{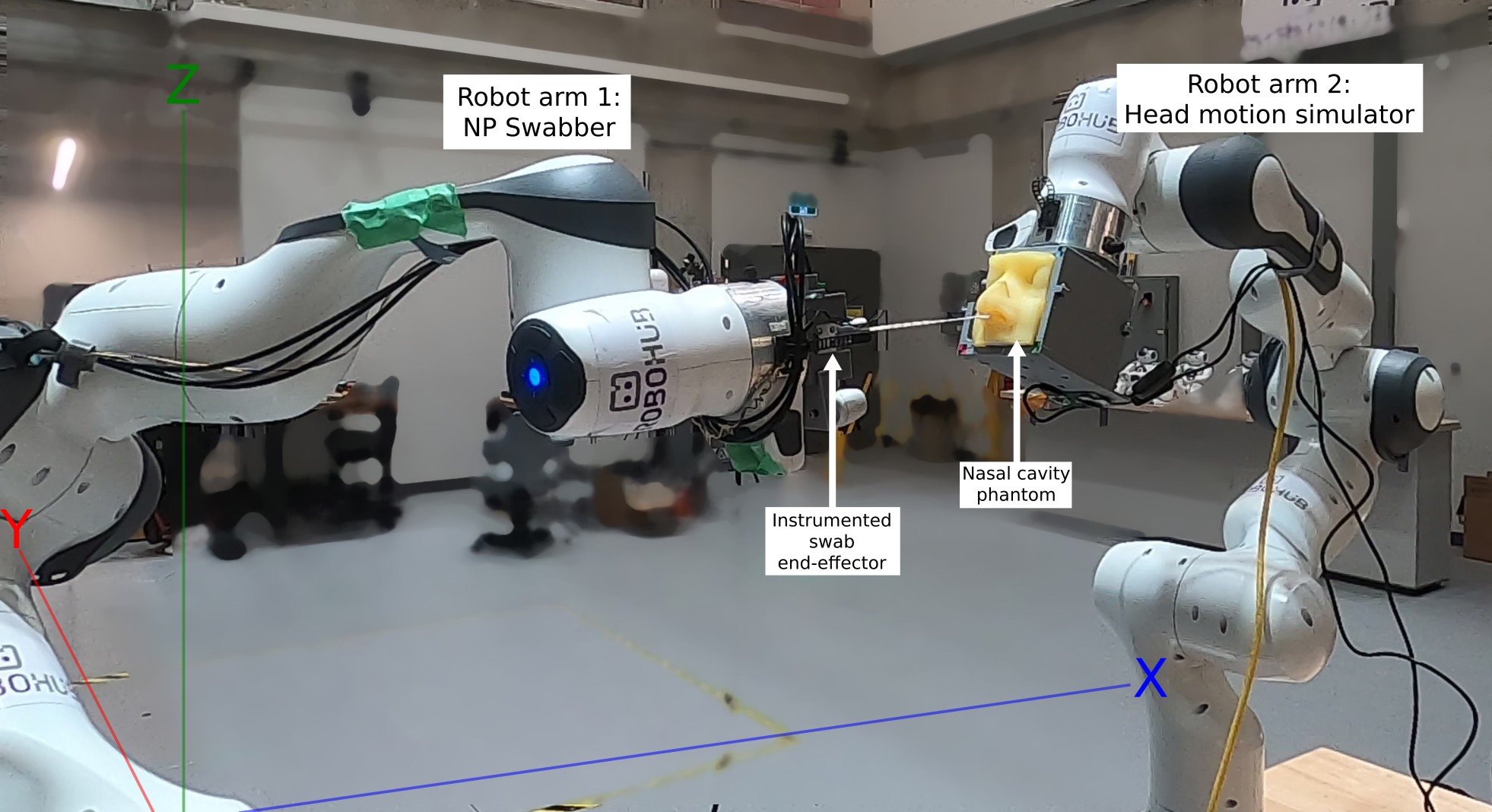}
  \caption[Overview of two robot arms interacting]{Figure showing the setup between the two robotic arms. One arm performs the NP swab test using the instrumented end-effector, while the second arm holds the nasal cavity phantom and simulates head motion.}
  \label{fig:int:overview}
\end{figure*}

    The application of autonomous robotics can greatly enhance clinical environments. Routine close-contact tasks place human healthcare workers (HCWs) at risk of direct infection from the patients they are treating, which is a significant workplace hazard. In addition, aging populations and the struggles in procuring an adequate workforce for the healthcare system are also factors that make applying robotics in these settings appealing. Robotics also provide a way to provide consistency in performing tasks that may otherwise suffer due to inconsistent training of HCWs (e.g., \cite{Hiebert2021-278}).
    The nasophraryngeal (NP) swab test is one example of a close-contact healthcare task that could benefit from automating with robots. The test involves inserting the swab through the nasal cavity in order to collect excretions from the nasopharynx at the exterior. The samples can be subjected to biochemical processes that can be used to test various types of respiratory illnesses, including types of coronavirus, influenza, pneumonia, and rhinovirus~\cite{Leber2020-173}.
        The NP swab test is particularly interesting because it requires fine control levels of control and sensing when the nasal cavity is occluded from the exterior. As a result, the techniques used to accomplish this task with robotics could push the envelope to enable similar types of close-contact tasks. 
    
            Robotic applications for upper-respiratory tract swabbing has received increased attention because of the COVID-19 pandemic. Oropharyngeal swabbing has the swab enter through the open mouth to collect samples along the throat, and several composite systems have been proposed that rely on vision or force controlled systems. For example,     the LINGCAI-II~\cite{Wang2024-423} is an autonomous robotic platform that was designed by researchers to perform oropharyngeal swabbing. A manipulator arm positions a specialized end-effector to conduct the swab actions in the mouth that are guided with a deep learning computer vision pipeline and force feedback. The platform also contains components to handle loading swabs and sample storage. 
    Sun \textit{et al.}~\cite{Sun2023-420} created a platform that consists of a manipulator arm and a seperate specialized 3-axis actuator. The manipulator is responsible for loading swabs and storing samples, while the separate 3-axis actuator performs oropharyngeal swabbing, using inputs from an RGB-D camera and a force sensor.         Zhang \textit{et al.}~\cite{Zhang2021-215} took a soft-robotics approach and designed a rigid-flexible coupling manipulator to finely control contact points. The system was outfitted with a vision and force controlled system
to perform oropharyngeal swab tests through the open mouth. The apparatus was also eventually mounted onto a UR5 collaborative robot to allow for a wider workspace~\cite{Chen2022-391}.
Swabs entering through the nasal cavity are also used for diagnosing respiratory illnesses. Shallow nasal swabs and swabs reaching the mid-turbinate are sometimes performed and are less invasive and easier to perform, but are less diagnostically sensitive~\cite{Pinninti2020-398}. In contrast, cultures from the nasopharynx are considered to be the gold-standard for diagnosing COVID-19 among other respiratory illnesses, but are more difficult to perform because the swab must travel to the posterior of the nasal cavity and both the swab and nasal cavity are visually occluded during the process.
Chen \textit{et al.}~\cite{Chen2022-313} created a hybrid rigid soft robot that is designed to mimic the dexterity of the hand and be compliant to measured force feedback. The robot is operated with teleoperation to eventually perform both nasopharyngeal and oropharyngeal swabbing.
Maeng \textit{et al.}~\cite{Maeng2022-314} designed a robot around a specialized force restriction mechanism that limits the maximum force the swab can apply during NP swab tests.  The swab is actuated through the nasal cavity using a remote centre of motion mechanism.
The robot for Shim and Seo~\cite{Shim2024-421} also used a remote centre of motion actuator to perform NP swabbing using teleoperation.
Zhang \textit{et al.}~\cite{Zhang2023-355} reported results from a wheeled dual-arm humanoid robot system that utilizes vision and a force-torque
sensing to perform NP swab tests.  Haddadin \textit{et al.}~\cite{Haddadin2024-390} used a Franka Emika research arm to autonomously perform swabbing in the nasal cavity and through the mouth within a number of human trials. However, it is unclear if their system actually reaches the nasopharynx because the reported displacement after it enters the swab enters nostril is far below the 9 to 10 cm needed to reach the nasopharynx~\cite{LimHyungsun2014-395}. As well, their setup utilizes a static fixation that the patients place their nose and mouth
against, and does not extend to adapting to variable poses from a freestanding
patient scenario.

 We take the perspective of using a non-specialized robotic platform for performing the test that could conceivably be applied to other close-contact health tasks. The system is built around using a collaborative manipulator equipped with an instrumented end-effector to hold a tool with an electromagnet, along with a camera and force sensing system to guide the control systems, which we show in Section \ref{sec:methods}. Further, we assume that patients would stand unrestrained during the swabbing process, meaning that the system must adjust to the pose of the head, and adjust for natural head motions. This approach better reflects the conditions under which HCWs performed tests in clinical environments. Also, it does not require maintaining an external fixture and can be a way to enhance patient comfort.
Consequently, we divide the NP swab test into two major phases: a pre-contact phase where the swab is aligned with respect to the patient's nostril and a contact phase where the swab is inside the nasal cavity.
 In our previous works, we have proposed solutions to the different subproblems pertaining to NP swab sample collection. First, we examined the question of finding the optimal trajectory to insert the swab through the nasal cavity, which we solved by leveraging a simulation and optimization approach~\cite{Lee2022-309}. Then we designed a visual servo system that aligns the swab next to the patient using information from an eye-in-hand RGB-D camera, with the pose of the face being interpreted with a deep learning model~\cite{Lee2024-437}. As well, we showed the feasibility of compliant torque control law to perform the contact phase of the test to demonstrate the ability for the system to adjust for misalignment~\cite{Lee2024-436}.

 The contributions of this paper are two-fold. First, we build on and integrate of the components from our previous works into a cohesive system that can execute the both the pre-contact and the contact stages of the NP swab test on hardware, which we describe with Sections \ref{sec:methods}, \ref{sec:mod_visual_servo}, \ref{sec:optimal_path}, \ref{sec:compliant_controller}, \ref{sec:term_obs}, and \ref{sec:safety}. Second, we conduct in-depth experiments to evaluate the suitability of the combined system towards safely executing the test. To facilitate this, we utilize a second robotic arm platform to hold a nasal cavity phantom that emulates natural head motions, described in Section \ref{sec:head_model}, which we use to evaluate the robustness of the swab system under different conditions.  Fig. \ref{fig:int:overview} provides an overview of how the two robots operate with respect to one another. These experiments, shown in \ref{sec:int_experiments}, were used to identify controller configurations that were able to effectively perform the test even with heavy head motion. Finally, Section \ref{sec:int:discussion} provides and analysis of the results, while \ref{sec:int:conclusion} provides concluding remarks and outlooks. To accompany this paper, a video is attached (swab\_vid.mp4) in the supplementary files that provides an overview of the proposed system and experiments.

\section{System overview}
\label{sec:methods}
\begin{figure}
  \includegraphics[width=\linewidth]{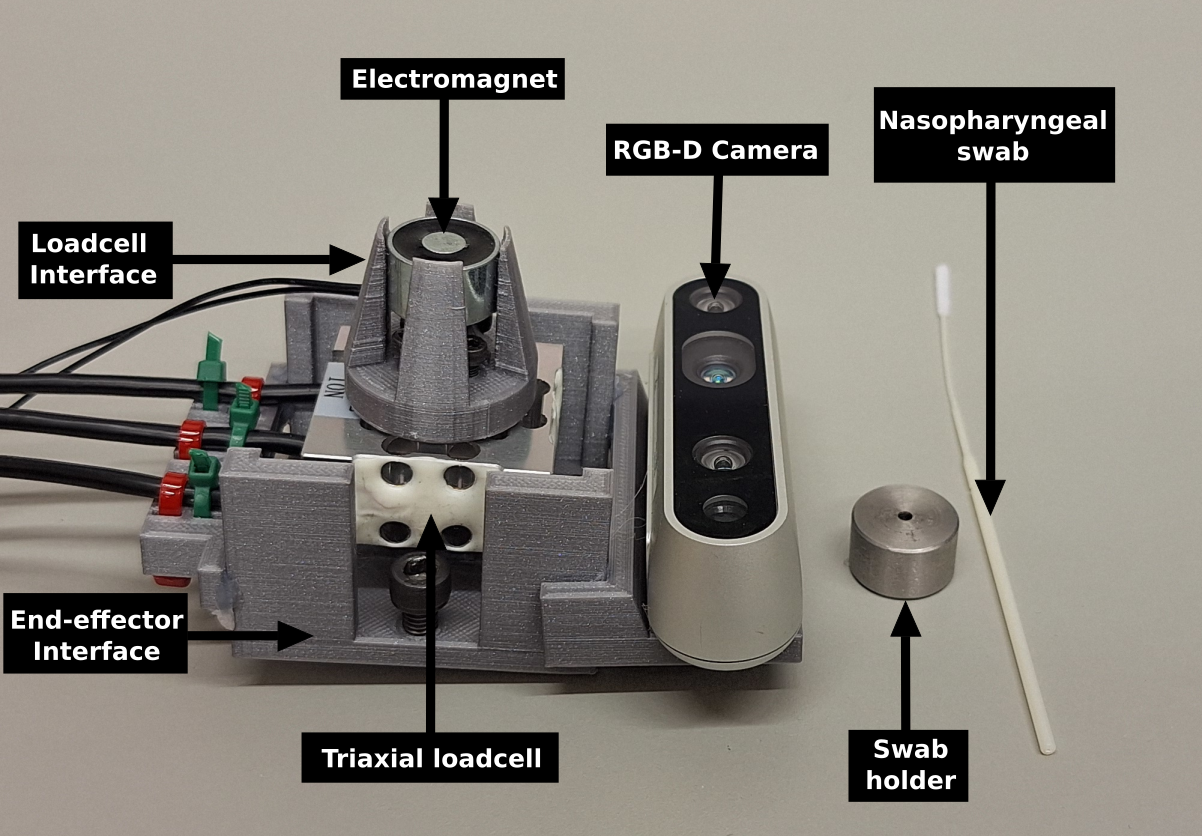}
  \includegraphics[width=\linewidth]{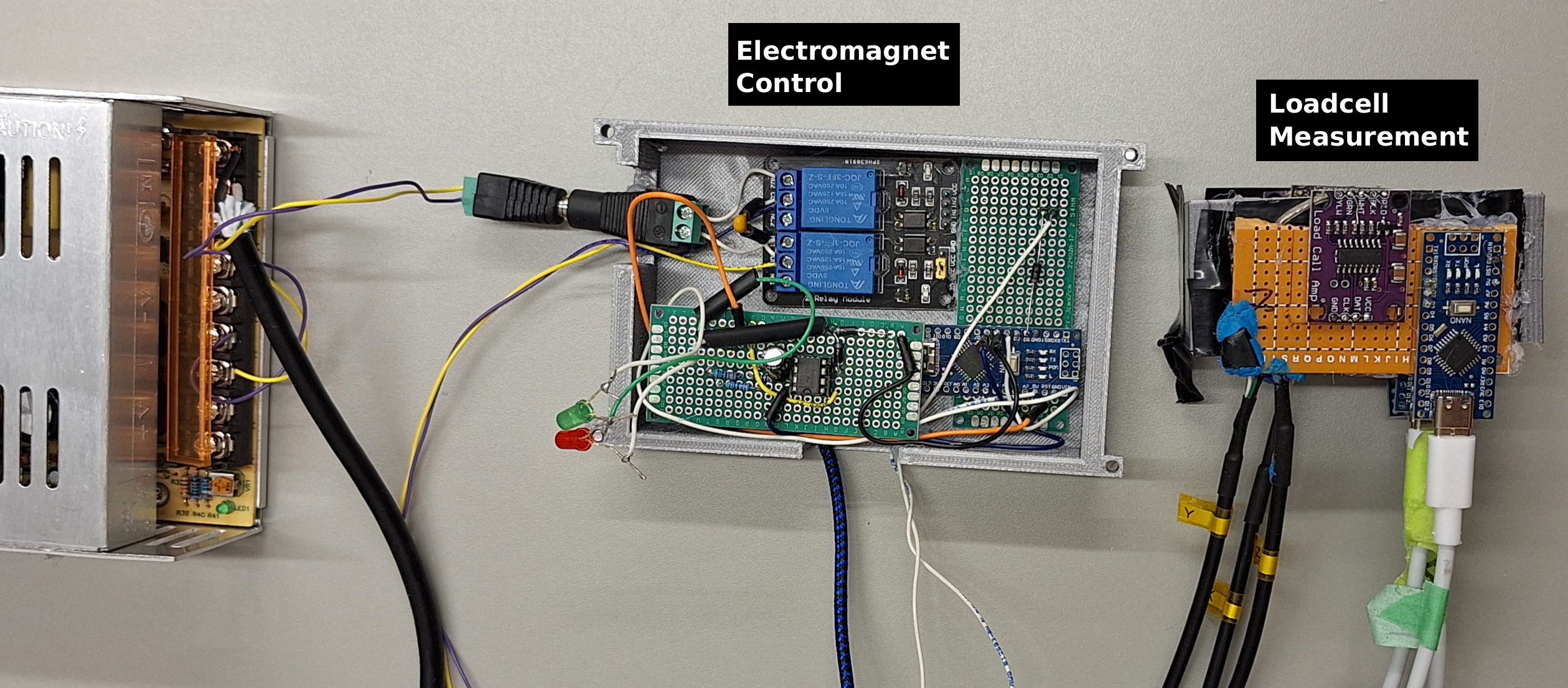}
  \caption{Top: Image showing the instrumented swab end-effector, which features a tri-axial loadcell, an RGB-D
    camera, and a electromagnet system to attach and detach the NP swab. Bottom: supporting
    electronics for controlling power to the electromagnet and ADC circuitry to measure forces on
    the loadcell.}
  \label{fig:end-effector}
\end{figure}

The focus of our work is to develop and evaluate the full NP swab test and
evaluate its performance under variable conditions. The hardware setup consists of
two Franka Emika arms that face one another during experiments.
    The first arm acts as the platform that performs the
swabbing, as shown in Fig. \ref{fig:end-effector},
and the second arm is the platform that simulates a human patient that randomly moves its head during the procedure, as will be elaborated later in Section \ref{sec:head_model}.
The end-effector attached to the first arm consists of a custom interface that was 3D printed using PLA. Due to the insensitivity of the onboard torque sensors of the Franka arm, the end-effector contains a three-axis, 10 N capacity loadcell on
the other end to support fine compliant control. Attached to the loadcell is a second interface that
holds an electromagnet rated for 30 N of attraction. The electromagnet holds a small 19.05 mm
diameter 416 stainless steel cylinder piece that was milled to fit a swab. Supporting circuitry is used
to allow the controller or human to disconnect the electromagnet from its 24V power supply unit in case of an abort scenario. An unpolarized capacitor is used to connected to the terminals of the electromagnet to counteract residual magnetism.
An Intel Realsense 435i depth camera is attached to the end-effector, facing slightly
inwards, so the swab is in frame in order to support visual servo operations.

\begin{figure*}
  \includegraphics[width=\linewidth]{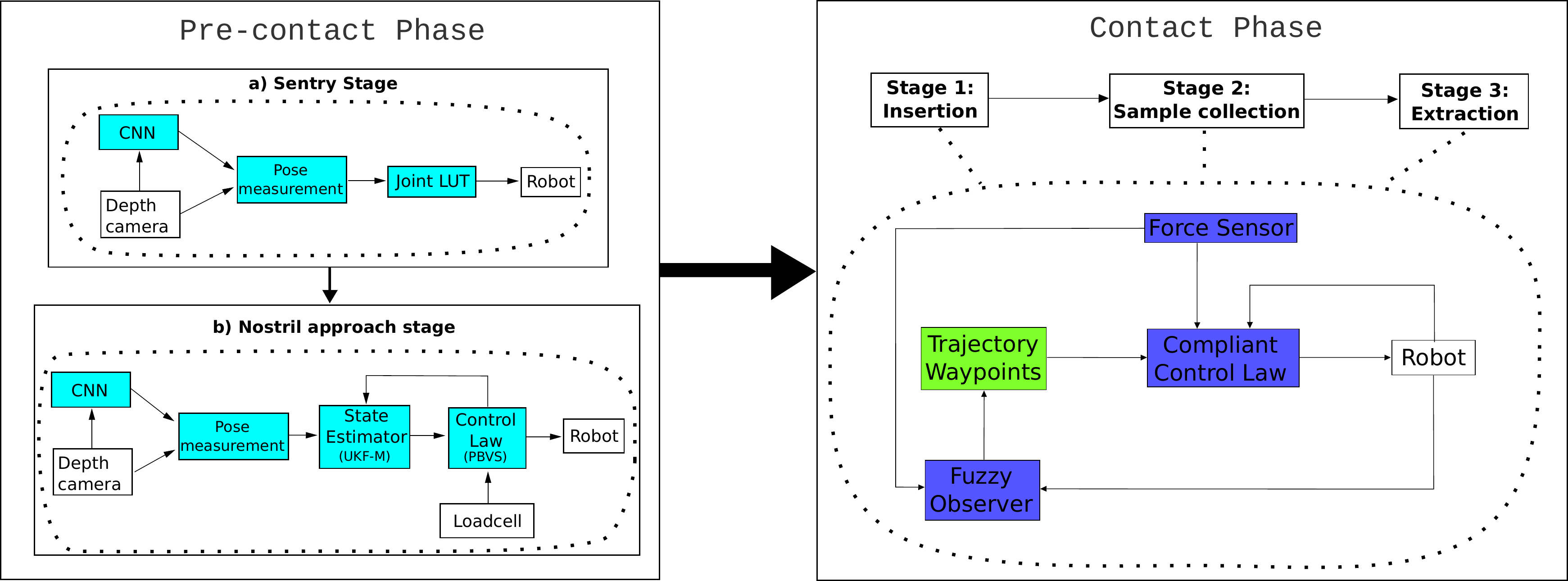}
  \caption[System diagram and staging]{Diagram of the stages and proposed system for a robot executing the NP swab test. The test is divided into a pre-contact
  phase and a contact-phase. The pre-contact phase is responsible for aligning the swab next to the
  nose using visual information. A sentry stage locates the face within the arm's workspace and
  moves to a closer joint configuration, while the nostril approach phase uses visual servo to
  place the swab next to the nostril. The contact phase has an insertion stage that moves the swab
  through the nasal cavity, the sample collection stage that holds and rotates the swab at the
  nasopharynx, and an extraction phase where the swab is removed from the nasal cavity.}
\label{fig:task-diagram}
\end{figure*}

As shown in Fig. \ref{fig:task-diagram} the swabbing task can largely be divided into a pre-contact phase
and contact phase. The pre-contact phase uses
the camera to guide movement from its sentry position to a suitable joint configuration close to the
face. Then the visual servo system brings the swab towards the nostril with the correct position and
orientation.
The contact phase is further divided into three substages, as shown in Fig. \ref{fig:task-diagram}.  The first substage inserts the swab
through the nasal cavity. Once the swab reaches the nasopharynx, stage 2 begins, where the swab stays at the nasopharynx and rotates ($\pm$
$45^\circ$) for 15 seconds to emulate collecting a sample. Stage 3 extracts the swab from the nasal cavity. The baseline
trajectory followed during stage 1 and in reverse for stage 3 is derived via an optimization problem described in Section
\ref{sec:optimal_path}. The contact phase uses a compliant joint velocity controller, which will be described in Section \ref{sec:compliant_controller}, that follows a prescribed baseline trajectory while reacting to forces applied to the swab.  The transition between stage 1 and 2 hinges on detecting when the swab reaches the nasopharynx, which is handled by a fuzzy logic system that is defined in Section \ref{sec:term_obs}. Finally, 
we also implemented a safety monitor to facilitate aborting the procedure in
case safety limits are violated (Section \ref{sec:safety}).

\section{Modified visual servo system}
\label{sec:mod_visual_servo}

In our previous work~\cite{Lee2024-437}, we designed a visual servo system that was able consistently bring the swab at the nostril of
human participants with a specified pose. Briefly summarized, a joint lookup table is employed to move the
end-effector in proximity of the face during the sentry phase, while avoiding joint limits. A pipeline surrounding the 3DDFA\_V2 CNN~\cite{Zhu2016-121}
produces pose estimates of the face, which are then fed into a UKF-M state estimator coupled with a PBVS control loop.
We reuse this system for our phantom experiments, but add a few modifications; the 3DDFA\_V2 CNN is substituted, the gain for the PBVS is modified, and a force feedback term is placed in the control loop.

The visual system described in this section estimates the 3D position and orientation of the phantom's nostrils. This approach replaces the 3DDFA\_V2 CNN~\cite{Zhu2016-121} used in our previous work~\cite{Lee2024-437}, which was trained to recognize anthropomorphic faces - a characteristic the nasal cavity phantom lacks.
The system relies upon detecting the five fiducial markers implanted onto the 3D printed fixture, as later shown in Fig. \ref{fig:head_rig}.
The fiducials are printed patterns of four coloured squares with the intersecting corner being the reference point. Thus, if the visual system can determine the 3D position of at least three markers, the pose of the target nostril can be
determined with standard Euclidean rigid registration.

While the markers were designed to be easy to spot, their detection is still subject to lighting
conditions that make conventional computer vision techniques less robust. As a result, we used the
following pipeline to measure the pose of the phantom. First, we trained a Faster R-CNN~\cite{Ren2017-399}
convolutional neural network, using a Resnet50~\cite{He2015-402} backbone, to find the bounding box of the five markers and the phantom
within RGB images. The network was pretrained on the Coco~\cite{Lin2014-401} dataset and implemented with the MMdetection framework~\cite{Chen2019-400}.
The concept of retraining a previously trained model onto a new task is known as transfer learning~\cite{He2015-402}, and it enabled us to retrain the detector to detect fiducials with relatively few training examples compared to training it from scratch.
The Faster R-CNN finds the bounding box of the marker, and the pixel position is further refined by finding the intersecting corner of the
four squares by applying Harris corner detection~\cite{Sagonas2013-375}. The response is weighted with a Gaussian prior at the center of the box to prevent
the detection being skewed by noise outside the marker.
The registered depth channel takes the pixel location of
the corner and converts it into a 3D coordinate. Once the coordinates of each of the fiducials are recovered, the pose of the desired nostril can be recovered
through least squares rigid registration with respect to their corresponding positions from the schematics.
The measurements are passed through an unscented Kalman filter on manifolds (UKF-M)
\cite{Brossard2020-191}, which is allows a) noise filtering from the pose estimation, b) 
motion integration from the end-effector into state estimates, and c) maintaining state when measurements are lost from
depth acquisition or occlusion.

\begin{figure}
  \includegraphics[width=\linewidth]{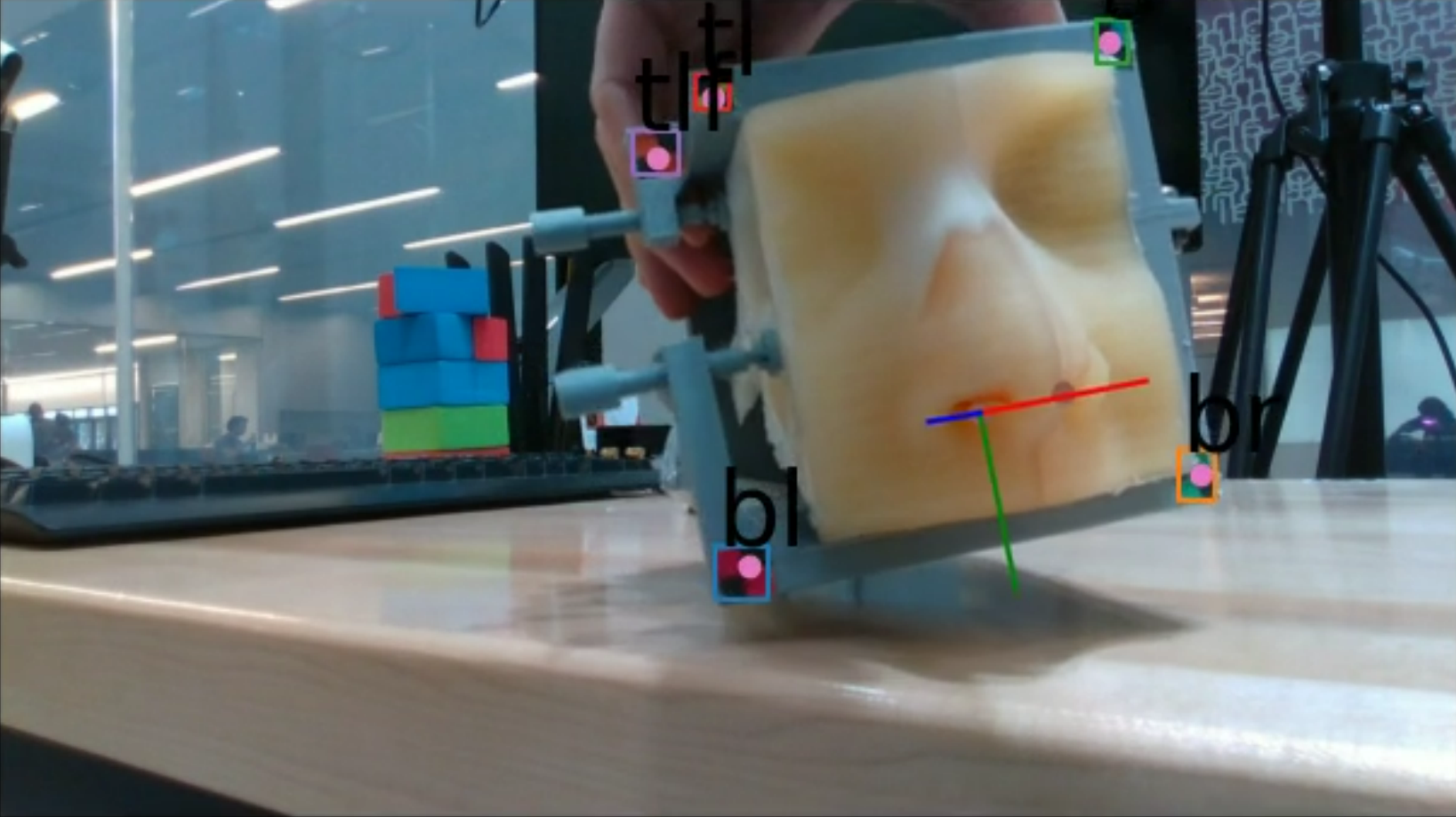}
  \caption[Pose estimation on phantom fixture]{Visualization of the pose estimation system adapted for the phantom fixture. The location of at least three of the fiducials leads to the identification of the nostril pose (red-green-blue axes).}
  \label{fig:non_anthro_pose}
\end{figure}

A pose based visual servo control law computes camera velocity to reach a desired target
\begin{equation}
  \mathbf{v}_c = -\lambda(||\mathbf{s}||) \hat{\mathbf{L}}^{\dagger} \mathbf{s},
\end{equation}
where $\mathbf{s}$ is the pose difference vector between the target and the swab tip and $\hat{\mathbf{L}}^{\dagger}$ is the
pseudo-inverse of the interaction
matrix $L$ (see \cite{Lee2024-437}), that maps the observed error to the camera's velocity frame. 
To allow the visual servo to converge with potentially large amounts of head motion, we also
implement the adaptive gain, $\lambda(||\mathbf{s}||)$, described by Kermorgant and Chaumette~\cite{Kermorgant2014-365}.
We chose parameters $\lambda(0) = 2.5$, $\lambda'(0) = 30$, and $\lambda(\infty)=0.5$, which enabled the visual servo to
converge quickly when there is high head motion. 

One additional component we added to the visual servo system is a force feedback term. Especially when the head is undergoing heavy motion, there is a chance the swab can make contact with another part of the face, such as the tip of the
nose, while it approaches the nostril. To avoid the swab from getting obstructed, we utilize the force readings from the loadcell make a
repelling motion away from the point of contact:
\begin{equation}
  \begin{aligned}
  \dot{\mathbf{f}}_1 &= -\alpha_1 \mathbf{f}_1 + \alpha_1 \mathbf{F}\\
  \dot{\mathbf{f}}_2 &= -\alpha_2 \mathbf{f}_2 + \alpha_2 \mathbf{F}\\
  \dot{\mathbf{q}} &= {}^{c}\mathbf{J}_{ee}^{\dagger} (  \mathbf{v}_c + {}^c\mathbf{M}_{ee} \bm{\Lambda}(\mathbf{f}_1 - \mathbf{f}_2)),
  \end{aligned}
\end{equation}
${}^{c}\mathbf{J}_{ee}$ is the robot Jacobian of the end-effector in the camera's frame, ${}^c\mathbf{M}_{ee}$ is
the transform from the end-effector to the camera frame, $\alpha_1 = 0.75$, $\alpha_2=0.6$, and $\bm{\Lambda} = \text{diag}(0.2,0.2,0.9)$ N$^{-1}$m/s.
The key purpose of the force term is to respond to a sudden change in force from contact events, but to also ignore smaller steady state
errors, e.g. from gravity, that would inhibit the controller from reaching the intended visual target.
To do this we form a band-pass filter by following the commonly known strategy of taking the difference of two low pass filters~\cite{Smith1999-379}.
Thus, the response of $\mathbf{f}_1-\mathbf{f}_2$ will spike shortly after contact and asymptotically converge to zero.

\section{Trajectory optimization}
\label{sec:optimal_path}
Once the swab tip arrives at the nostril, the contact stage controller requires some nominal trajectory to move along that would enable it to reach the nasopharynx.
In our previous work \cite{Lee2022-309},
we developed a solution for this problem by creating a finite element simulation of the swab 
and finding the optimal linear path in terms of minimizing strain on the swab. We revisit this
because the experiments from previous work~\cite{Lee2024-437} revealed that it is necessary for the
swab to arrive at the nostril at an incline angle. As a result, we modify our previous approach by formulating a new
non-linear trajectory and using a new collision model for the nasal cavity.  The trajectory parameterization has
changed to start at a $30^\circ$ angle pitched upwards from the transverse plane (increased from 0.2 radians~\cite{Lee2024-437} to allow for easier insertion) and parallel to the sagittal
plane at the centre of the nostril. A 3D vector ($\chi, e_1, e_2$) characterizes the trajectory.
The trajectory is divided into three parts.
\textbf{i)} the swab moves a distance of $\chi$  from the starting point at the initial angle; \textbf{ii)} the swab rotates about the tip until it points at a target position on an ellipse at the nasopharynx  (parameterized by $e_1$,
$e_2$); \textbf{iii)} the swab moves along a straight line towards the target for 150 mm, although the
simulation terminates once the swab intersects with the
ellipse, which is a distance of about 80 mm from the beginning of part iii). As before, the vector is optimized to minimize strain energy using COBYLA \cite{Powell1994-240}.
The new collision model is based off of the nasal cavity phantom from Sanan{\`e}s \textit{et al.}~\cite{Sananes2020-91}, which has the nose
overhanging the nostril.

\begin{figure}
  \includegraphics[width=\linewidth]{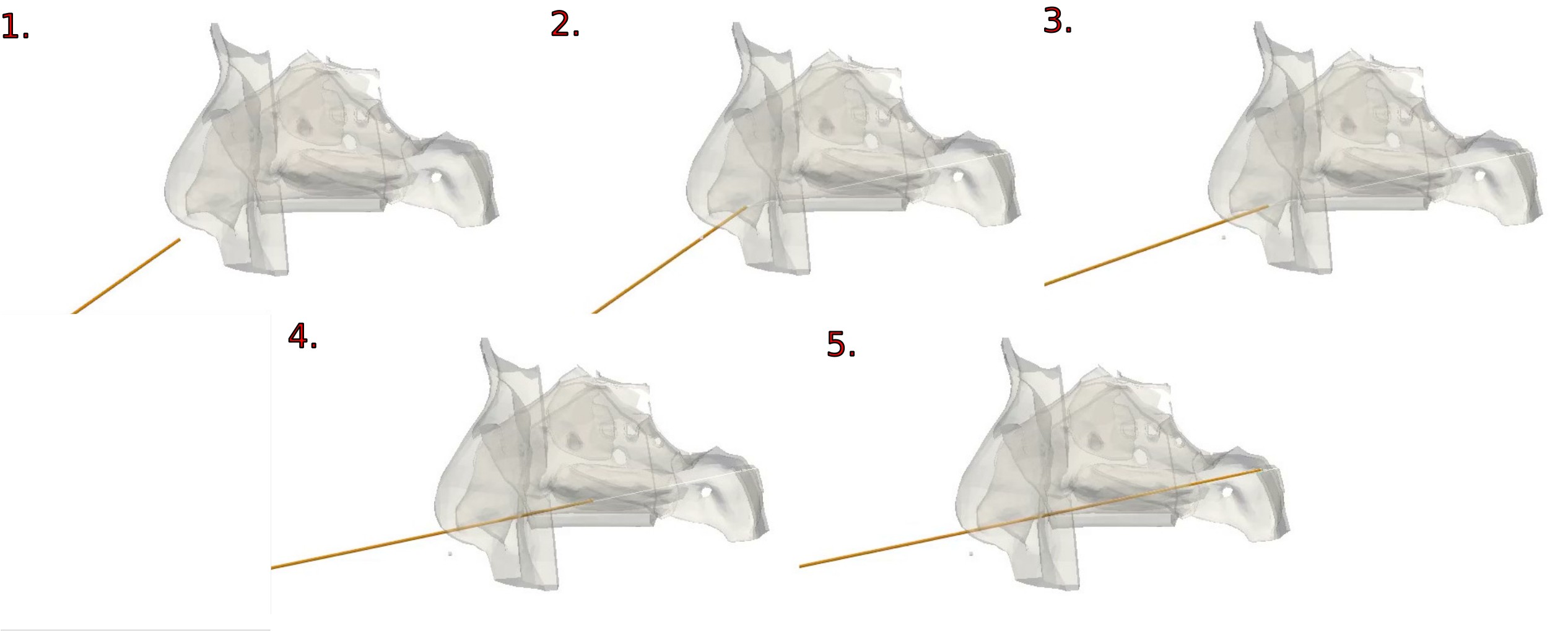}
  \caption[Optimized insertion trajectory through nasal cavity]{Sequence of the optimized trajectory as it travels through the nasal cavity volume. Notice
    how this trajectory begins at an incline angle and then rotates shortly after entering to point
    at the nasopharynx.}
  \label{fig:optimized-trajectory}
\end{figure}

Fig. \ref{fig:optimized-trajectory} shows a sequence of the optimized trajectory within the
simulation, with $\chi=17.3$ mm and a change in angle during part iii) of $25.6^\circ$ pitch and $-3.9^\circ$ yaw. The trajectory is angled towards the septum to avoid the inferior turbinate and also angled slightly
upwards to avoid the strain of becoming wedged on the nostril. The upwards angle appears to be the main difference from
the previous result~\cite{Lee2022-309}, as the optimization balances the trade-off between avoiding wedging on the nostril and avoiding the inferior turbinate at the back of the nasal cavity.

The selected trajectory is prepared for run-time use by dividing it into a series of Cartesian
waypoints, indexed by time. The waypoints are smoothed with triangular ramped acceleration
for the beginnings and ends of parts i), ii), and iii) to prevent strain on the hardware. Overall the
entire trajectory would take 20 seconds, but 
in reality the real trajectories took less time because the tip would reach the nasopharynx (see
Section \ref{sec:term_obs}) before the remaining distance in part iii) could play out.
During runtime, the waypoints are transformed to the end-effector's frame of reference, then the
target is interpolated using a linear spline with respect to time.
Similar to our previous work \cite{Lee2024-436}, the rate of the
target trajectory, $l$, is adjusted depending on the level of axial force, $f_z$, by using a sigmoid function
\begin{equation}
  \frac{dl}{dt} = 1 - \sigma ( s_l(f_z - \bar{f}_l)), 
  \label{eq:slowdown_zaxis}
\end{equation}
with $s_l=60 \text{ N}^{-1}$, $\bar{f}_l = 0.5$ N.
 This means that when there is little axial
force, the trajectory proceeds at regular speed, but if high axial forces are observed, the
trajectory slows down to give the controller time to respond to the disturbance.

\section{Compliant contact stage controller}
\label{sec:compliant_controller}
Our previous work \cite{Lee2024-436} showed our initial attempt at applying a compliant controller to insert the swab by using a torque control law.
  The experiments showed that force input is necessary to correct for misalignments on a stationary phantom, and that a system unguided by force feedback has much higher failure rates.
  However, continuing to operate the arm in torque control mode is undesirable for several reasons.
The Franka Emika API requires sending commands to an external control box, which creates non-negligible networking latency that hamper the responsiveness of a torque controller operating under a second order differential system.
 Model inaccuracies and uncompensated friction also hinder
performance. Ultimately, higher precision is needed for a moving head than what is feasible in torque control mode. As a result, we
switched to a velocity control law to execute the contact phase.

While a torque compliant control law can be implemented by projecting the scaled forces measured on the end-effector to torque on the
joints, using forces in a velocity control law requires a different strategy. As force acts as a second-order
quantity in terms of the dynamics, a velocity controlled system needs to integrate force measurements to apply them as a first order term.
Like in  previous work~\cite{Lee2024-436} we apply a low pass filter to integrate the raw force measurements, $\mathbf{F}$, to the lower order quantity $\mathbf{f}$: 
\begin{equation}
  \dot{\mathbf{f}} = -\alpha \mathbf{f} + \alpha \mathbf{F},
  \label{eq:filter}
\end{equation}
with $\alpha$ as the decay rate of the exponential moving
average filter.
We reduced $\alpha$ to $0.5$ to adjust for its usage in velocity control.
Then we can add this term into our control law as 
\begin{equation}
  \dot{\mathbf{q}} = \mathbf{J}^{\dagger}(\mathbf{K}(\overline{\mathbf{e}} - \mathbf{e}) - \mathbf{W}\bm{\Lambda} (\overline{\mathbf{f}}-\mathbf{f})),
  \label{eq:vel_control_with_f}
\end{equation}
where $\mathbf{e}$ is the current end-effector pose, $\overline{\mathbf{e}}$ is the target pose, $\mathbf{J} \in \mathbb{R}^{6 \times 7}$ is the robot Jacobian, and $\mathbf{K}=\text{diag}(5,5,5,2,2,2)$ $\text{s}^{-1}$ is the diagonal
Cartesian pose gain matrix. The force gain matrix is $\bm{\Lambda} \in \mathbb{R}^{3\times 3}$,
while \[\mathbf{W} = \begin{bmatrix} 1 & 0 & 0\\0 &1&0\\0&0&-1\\0&1&0\\-1&0&0\\0&0&0\end{bmatrix}\] acts as a
projection matrix mapping the 3D force measurement to the 6D velocity axes. Notice that the fourth and fifth rows invoke some rotational motion
to account for the displacement between the swab and the end-effector.
The target force is $\overline{\mathbf{f}}$, which is held to $\overline{\mathbf{f}}=[0,0,0]^\top$ N for stages 1 and 3, and $\overline{\mathbf{f}}=[0,0,0.4]^\top$ N for stage 2, when the swab is held against the nasopharynx.
Given the choice of $\mathbf{K}$ and $\alpha$, the remaining question is how to choose an appropriate force gain $\bm{\Lambda}$. Ideally $\bm{\Lambda}$ should be chosen so that it is high enough that is responsive to input forces but
low enough so that chattering and instability do not occur. Rather than relying on trial-and-error,
a more elegant way of gain tuning can be found by
examining the mechanical properties of the swab.

\begin{figure}
  \includegraphics[width=\linewidth]{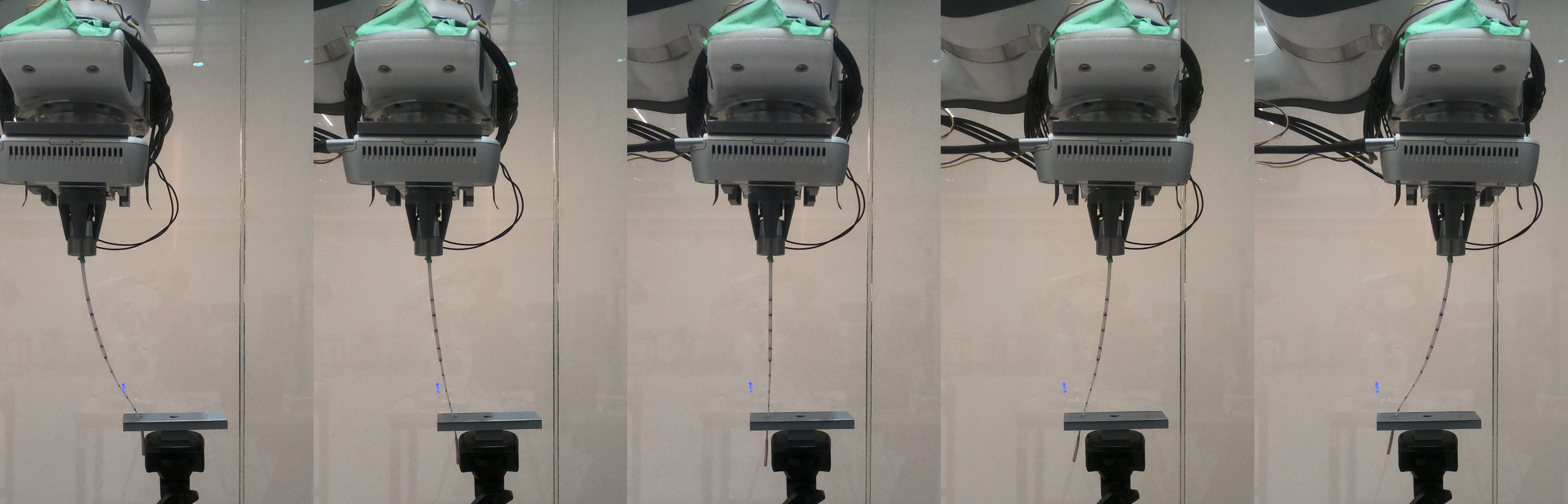}
  \caption[Swab held in fixture as end-effector moves in a transverse direction]{Swab held in fixture as end-effector moves in a transverse direction in order to evaluate if the swab exhibits linear elastic properties.}
  \label{fig:swab_deform}
\end{figure}

\begin{figure}
  \includegraphics[width=\linewidth]{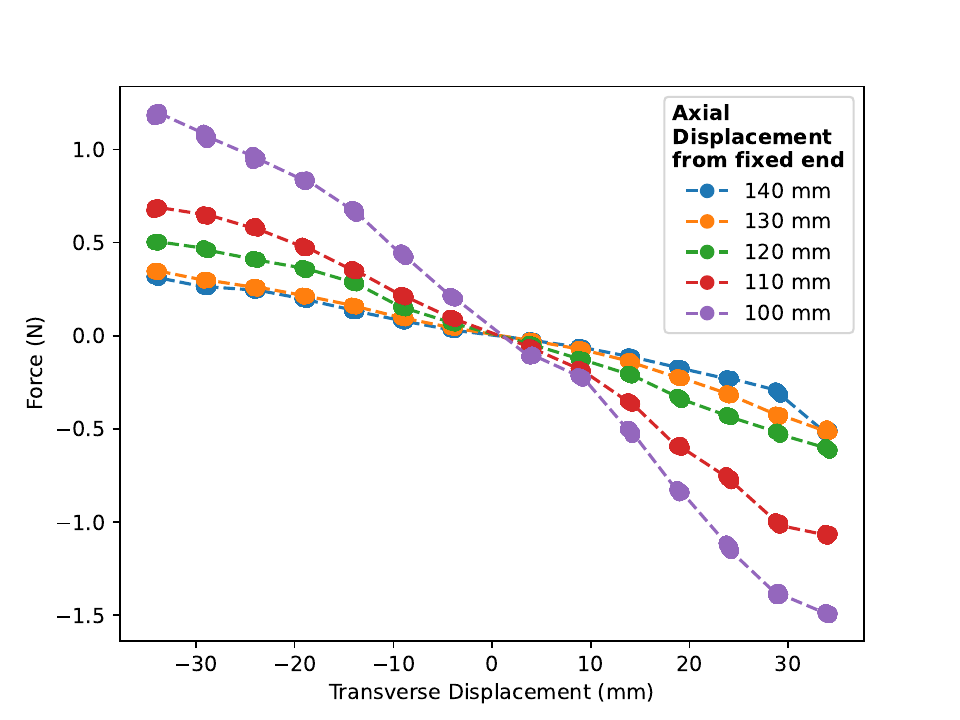}
  \caption[Transverse displacement vs. force]{Graph showing the relationship between displacement and force for different
    holding points within the fixture. For the range of motion we expect on the swab during the NP test, the relationship appears linear.}
  \label{fig:force_disp_rel}
\end{figure}

Based on the assumptions that the swab obeys the mechanics of an Euler-Bernoulli elastic cantilever beam~\cite{Lee2022-309},
we expect a linear relationship between any transverse displacement of the swab and the magnitude of force applied along that transverse axis.
The swabs used in these experiments are constructed from ABS plastic and so it is helpful to know if
the shaft will remain in the elastic region for the expected range of motion and also to uncover the
associated stiffness parameters.
Hence, we conducted an
experiment where the arm held the swab vertically in a fixture,
as shown in Fig. \ref{fig:swab_deform}, and the arm periodically  moved in a transverse direction while measuring
the forces. Fig. \ref{fig:force_disp_rel} shows the relationship between displacement and measured force. There appears to be
a linear relationship at multiple points along the shaft within the expected range of motion for the swab, with the average $R^2$ correlation coefficients being 0.983.

If we examine just a transverse axis, this linear elastic relationship means there will be a restoring force in the opposite
direction that could be measured at the fixed end
\begin{equation}
  F = -\nu \Delta x,
  \label{eq:linear_elastic}
\end{equation}
where $\Delta x$ is the transverse deformation and $\nu$ is the stiffness variable, dependent on the material properties of the swab, that will increase with loads applied further away from
the fixed end. Suppose we simplify the force and displacement to a single axis, and assume that the
setpoint and equilibrium point of contact are set to 0. If we were to assume
that the external body remains still, we can substitute 
(\ref{eq:linear_elastic}) into (\ref{eq:filter}), which in the context of the control law
(\ref{eq:vel_control_with_f}) yields a linear differential equation 
\begin{equation}
  \begin{aligned}
    \dot{x} &= k(- x) + \lambda f\\
    \dot{f} &= -\alpha f + \alpha \nu x \\
  \end{aligned}
\end{equation}
that can be represented as a homogeneous system:
\begin{equation}
  \begin{aligned}
    \dot{\mathbf{z}} &= \mathbf{A}\mathbf{z} ,\\
    \mathbf{z} &= \begin{bmatrix}x\\ f\end{bmatrix}, \mathbf{A} = \begin{bmatrix} -k & \lambda\\\alpha \nu & -\alpha\end{bmatrix}.
  \end{aligned}
  \label{eq:homogeneous_ODE}
\end{equation}
Thus, if scalar $k$ and $\alpha$ are set and $\nu$ is known, then the scalar gain $\lambda$ can be chosen to
determine the response characteristics of the system. Theoretically, choosing $\lambda$ to
critically damp the system, where the eigenvalues of $\mathbf{A}$ have no imaginary components~\cite{AAstrom2008-272}, should ensure that oscillations do not occur
for the steady state
response. The $\lambda$ that causes critical damping is
\begin{equation}
  \lambda_{crit} = \frac{{(\alpha + k)}^2/4 - k\alpha}{\nu\alpha},
  \label{eq:crit-gain}
\end{equation}
which is found by manipulating the characteristic polynomial of $\mathbf{A}$. Although some liberty was taken in the assumptions in this model (non-moving
contact point, no interaction between axes or rotational components), we expect $\lambda_{crit}$ to be stable and a good
baseline value to start from for the transverse axes. We will also explore gains higher than this, in case the critically damped gain
responds too slowly to disturbances or the response is too low to pull the swab out from the contact.

According to Euler-Bernoulli beam theory \cite{Bauchau2009-190}, the stiffness, $\nu$, will scale by the inverse cube of the
length along the beam's axis, $L$, from the fixed end. While determining the material properties on this scale 
is not trivial, we can take measurements of stiffness at different grasping points along the shaft using the fixture shown in
Fig. \ref{fig:swab_deform}, and then fit an affine cubic function to compute the inverse stiffness $\nu^{-1}$
\begin{equation}
  \nu^{-1} = mL^3 + b.
  \label{eq:nu_fixture}
\end{equation}
To estimate $m$ and $b$, we placed the fixture to hold at points of $L$ between 95 and 145 mm. Note that the affine offset is added to this model due to the change in shaft thickness of the swab after the breakpoint at $L=$ 80 mm. For each $L$ we measured the
force-transverse displacement pairs at these points, using the linear relationship (\ref{eq:linear_elastic}) to find
$\nu$. Least squares regression was used to find $m=43.89$ and $b=-0.0193$ from the $L$, $\nu$ pairs.

\begin{figure}
  \includegraphics[width=\linewidth]{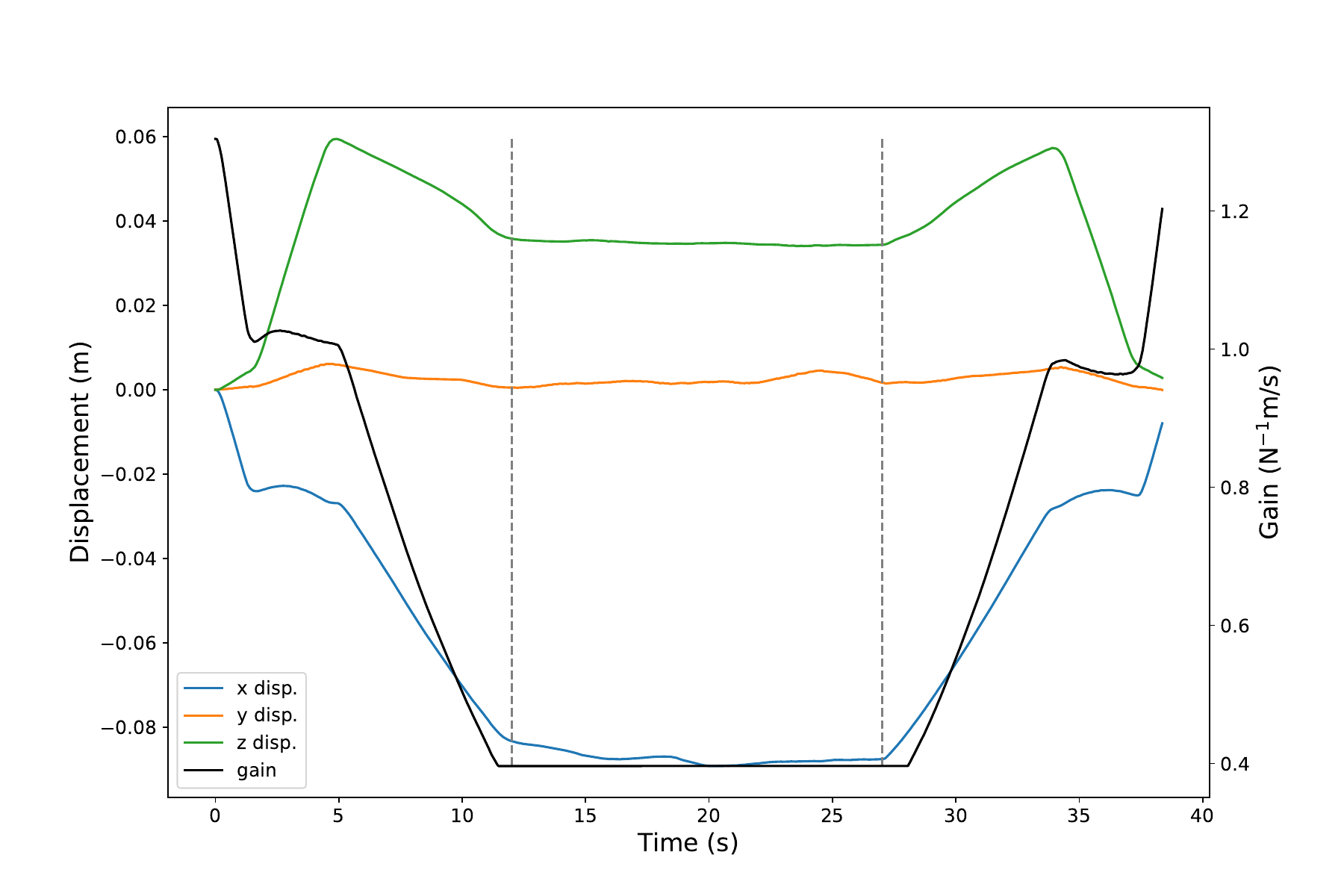}
  \caption[Transverse gain versus displacement]{Transverse gain computed for \textit{D1.0} for an example insertion. As the end-effector moves (mostly in the x-axis in the global frame), the gain scales down dynamically. The three stages of motion are divided by the vertical lines.}
  \label{fig:gain-tuning}
\end{figure}

Hence, we present two potential strategies for choosing suitable values of the transverse components of $\Lambda$. The first strategy is based around the hypothesis that the
majority of the contact occurs about the tip of the swab, where $\nu = 10.5$ N/m. As such, we consider three sets of
gains: the critically damped gain $\Lambda_{crit} = 0.880$ $\text{N}^{-1}\text{m}/\text{s}$ (\ref{eq:crit-gain}), as well as $1.5\Lambda_{crit}$ and $2.0\Lambda_{crit}$. These controller configurations will be identified as
\textit{S1.0}, \textit{S1.5}, and \textit{S2.0}. The second strategy
follows the hypothesis that the contact point moves further away from the tip of the swab is
inserted through the nasal 
cavity, and therefore will dynamically change $\Lambda$ as the displacement increases and $\nu$
increases. Given the tip
displacement as $\epsilon$, $\nu$ is interpolated using the cubic function (\ref{eq:nu_fixture}). To account for the
imperfect contact, we also scale the displacement by a factor of $\zeta=0.5$, so $L = L_{max} - \epsilon \zeta$, where $L_{max}=146$ mm. Again, we consider three sets of gains with respect to  the
critical gain (\ref{eq:crit-gain}) corresponding to $\nu( L_{max} - \epsilon \zeta)$ multiplied by factors of 1.0, 1.5, and 2.0, which will be identified as \textit{D1.0}, \textit{D1.5}, \textit{D2.0}. Figure \ref{fig:gain-tuning} shows a typical example of what the dynamic gain would be during the different stages of the insertion process.

\begin{figure}
  \begin{subfigure}{\linewidth}
    \includegraphics[width=\linewidth]{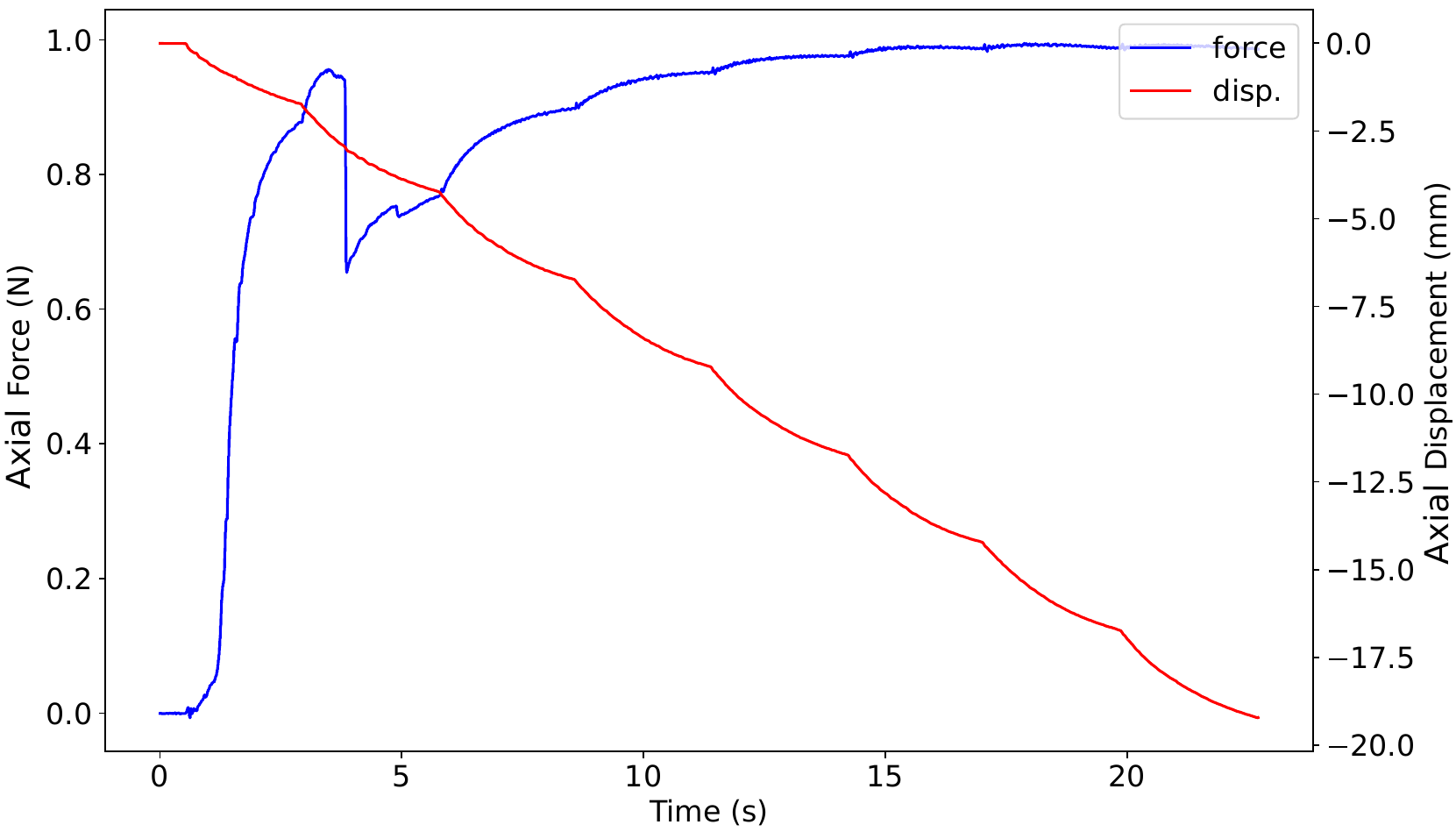}
    \caption{}
  \end{subfigure}
  \begin{subfigure} {\linewidth}
    \includegraphics[width=\linewidth]{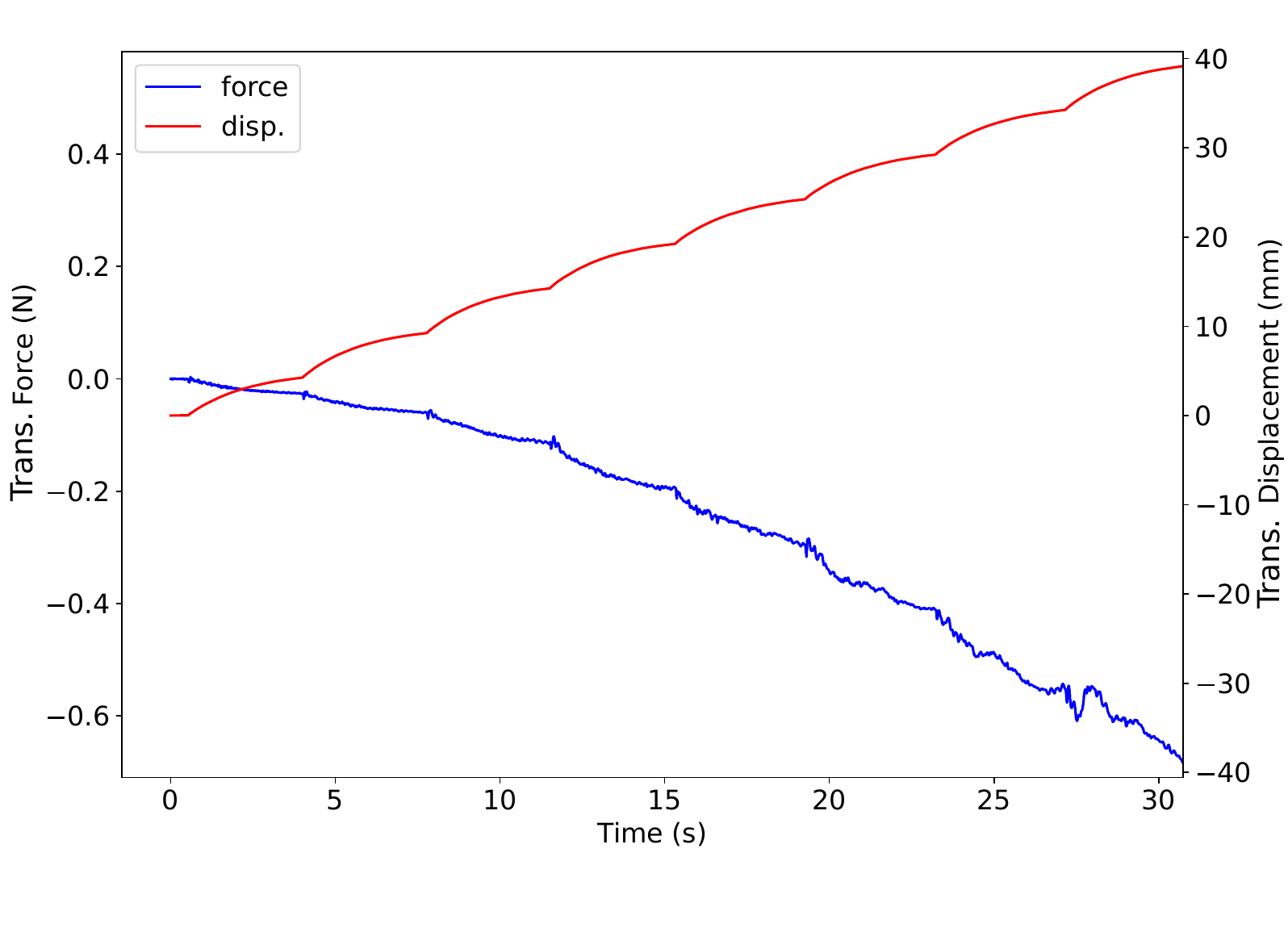}
    \caption{}
  \end{subfigure} 
  \caption[Attempt to estimate axial force model]{a) Attempt to recover axial force model of the swab. The measured forces behave like an impulse (around 1-2 seconds for the first
    mm of displacement) until the swab eventually buckles. b) For comparison, force-displacement relationship on a transverse axis, which
    exhibits a linear elastic relationship. }
  \label{fig:z-axial-force}
\end{figure}

The axial forces come from two sources. The first source occurs when the tip makes direct contact with the nasal cavity. Although the elastic beam theory would suggest that the force response should also act like a spring \cite{Bauchau2009-190}, in practice we observed that the response more like an impulse upon axial contacts. Although, once a certain threshold is reached (about 1 N of axial force), the swab begins to buckle and the axial force saturates (see Fig. \ref{fig:z-axial-force}), but this is less relevant for the level of force the controller will operate around. 
The second source of axial force comes from frictional forces as the swab moves forward during stage 1. When there are transverse forces applied to the beam, Coulomb friction will transmit a proportional axial force in the opposite direction of motion.
In the absence of a good model for these factors, we tuned the axial force based on the characteristics of stage 2 by taking the maximum gain before we started observing oscillations. This value is $0.051$ $\text{N}^{-1}\text{m}/\text{s}$.

\section{Nasopharynx contact observer}
\label{sec:term_obs}
A critical component to the proposed system is determining when the swab has reached the
nasopharynx. Since it is not possible to visually observe this event with an external camera, we must make the decision with two sources of
information: the displacement of the swab's tip and the total axial force returned. The otolaryngology literature estimates the distance
between the nostril to the nasopharynx to be between 90 and 100 mm, although this can vary depending on factors like age and gender
\cite{Liu2009-171} \cite{Keustermans2018-172} \cite{Reznikov1995-174} \cite{LimHyungsun2014-395}.
Liu \textit{et al.}~\cite{Liu2009-171} estimates the standard deviation of the length of the central portion of the nasal cavity at 11\%, so we can use this information to determine a lower bound of displacement to reach the nasopharynx at about 80 mm. The measured axial force is another good indicator for when the swab
reaches the nasopharynx because when
the swab's tip makes contact with the perpendicular nasopharynx, we observe a larger impulse in
force, as discussed in Section \ref{sec:compliant_controller}. 
Our study with healthcare workers~\cite{thesis} found a median peak force value at 0.86 N among 6 practitioners, while the study by Park \textit{et al.}~\cite{Park2021-301} found a peak force of 0.60 N for a single expert practitioner.

\begin{figure}
  \includegraphics[width=\linewidth]{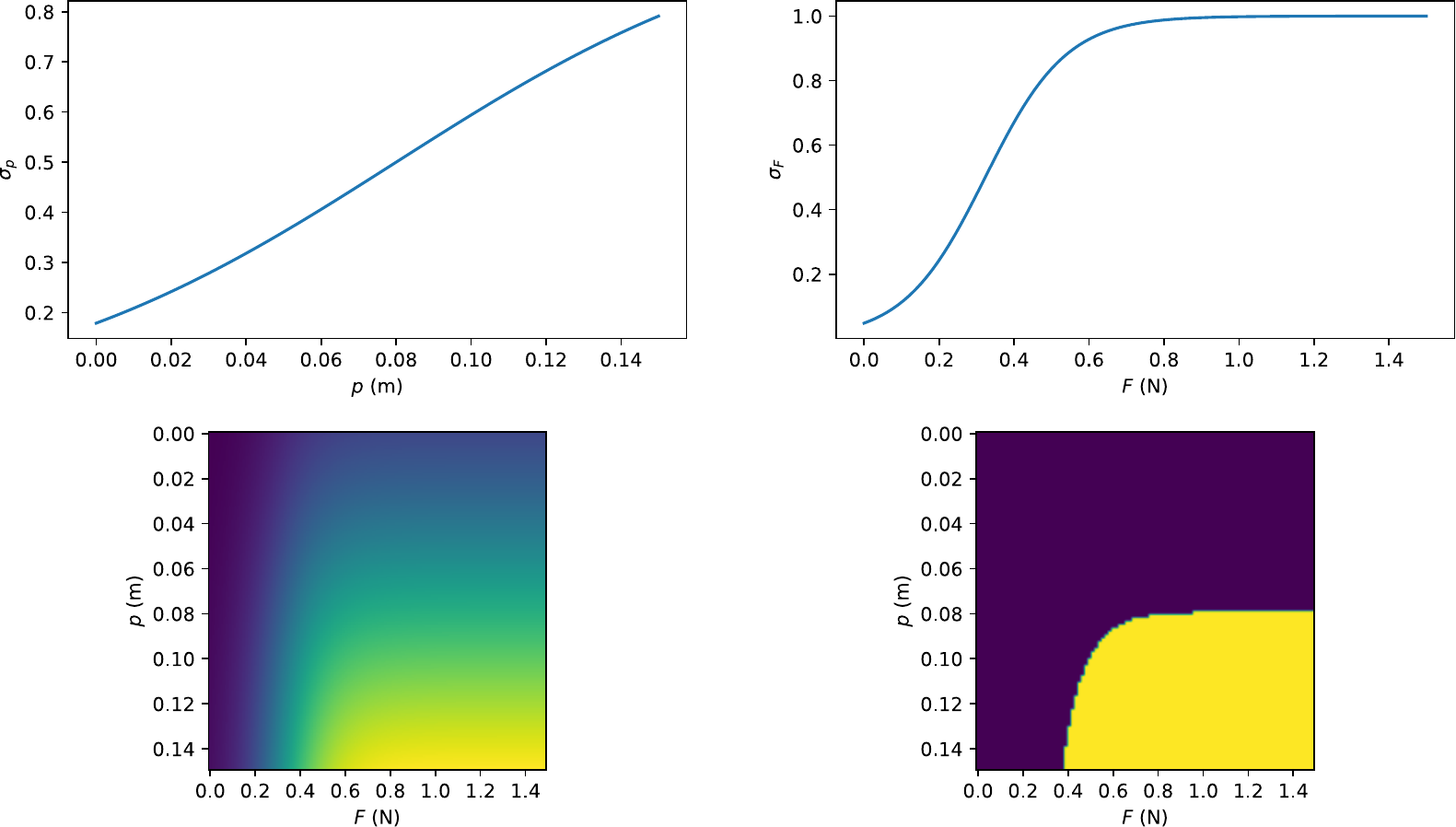}
  \caption[Fuzzy logic function responses]{Fuzzy logic sigmoid functions. The top left and top right graphs are the individual position and force sigmoids, whose outputs are unitless and bounded between 0 and 1. The bottom left graph shows the product between the two sigmoids, while the bottom right graph shows the corresponding decision boundary by thresholding at 0.5.}
  \label{fig:sigmoid}
\end{figure}

A robust observer needs to take into account multiple modalities. 
Displacement measurements alone are insufficient because the head may move during insertion, and nasal cavity dimensions vary across the population. Similarly, force measurements can fluctuate due to motion. 
Therefore, we propose a fuzzy logic system, similar to our previous work \cite{Lee2024-436}, to combine both force and displacement measurements for decision-making. This system employs a fuzzy AND (product) operation on two sigmoid functions, one for displacement ($p$) and another for axial force ($F_z$)
\begin{equation}\begin{aligned}
    \sigma_p &= \sigma(s_p(p - i_p)), \sigma_F = \sigma(s_F(F_z - i_F))\\
    o  &= \sigma_p  \sigma_F\\
    \sigma(x) &= \frac{(\tanh(x/2) + 1)}{2}.
  \end{aligned}\end{equation}
The outcome $o$ is unitless and bounded between 0 and 1. The decision rule for reaching the nasopharynx comes from thresholding $o$ at 0.5.
We use the equivalent hyperbolic tangent formulation of the sigmoid function as the function is implemented efficiently in the standard C++ math library. From
this system we have four parameters that need to be chosen: the scaling factors $s_p, s_F$ and intercepts
$i_p, i_F$ for the displacement and force sigmoids, respectively. The intercept term determines what
value the input must be for the sigmoid to saturate at 0.5, which means that the inputs must be
greater than the intercept in order for the decision to trigger. 
We tuned the intercepts at $i_p=0.08$ m, and $i_F = 0.323$ N based on the insights listed in the previous paragraph.
The scale parameters were placed at $s_p = 19.04$ ${\text{m}}^{-1}$, $s_F = 9.24$ $\text{N}^{-1}$ also based on the aforementioned supporting studies, and some initial experimentation with the phantom.
In contrast to our previous work~\cite{Lee2024-436} that chose parameters by trial and error, the information from the relevant literature inspired parameters that were much more accurate at detecting contact with the nasopharynx.
The shape of the two sigmoids, $o$, and the resulting decision barrier is shown in Fig. \ref{fig:sigmoid}.

\section{Safety observer}
\label{sec:safety}
\begin{figure}
  \includegraphics[width=\linewidth]{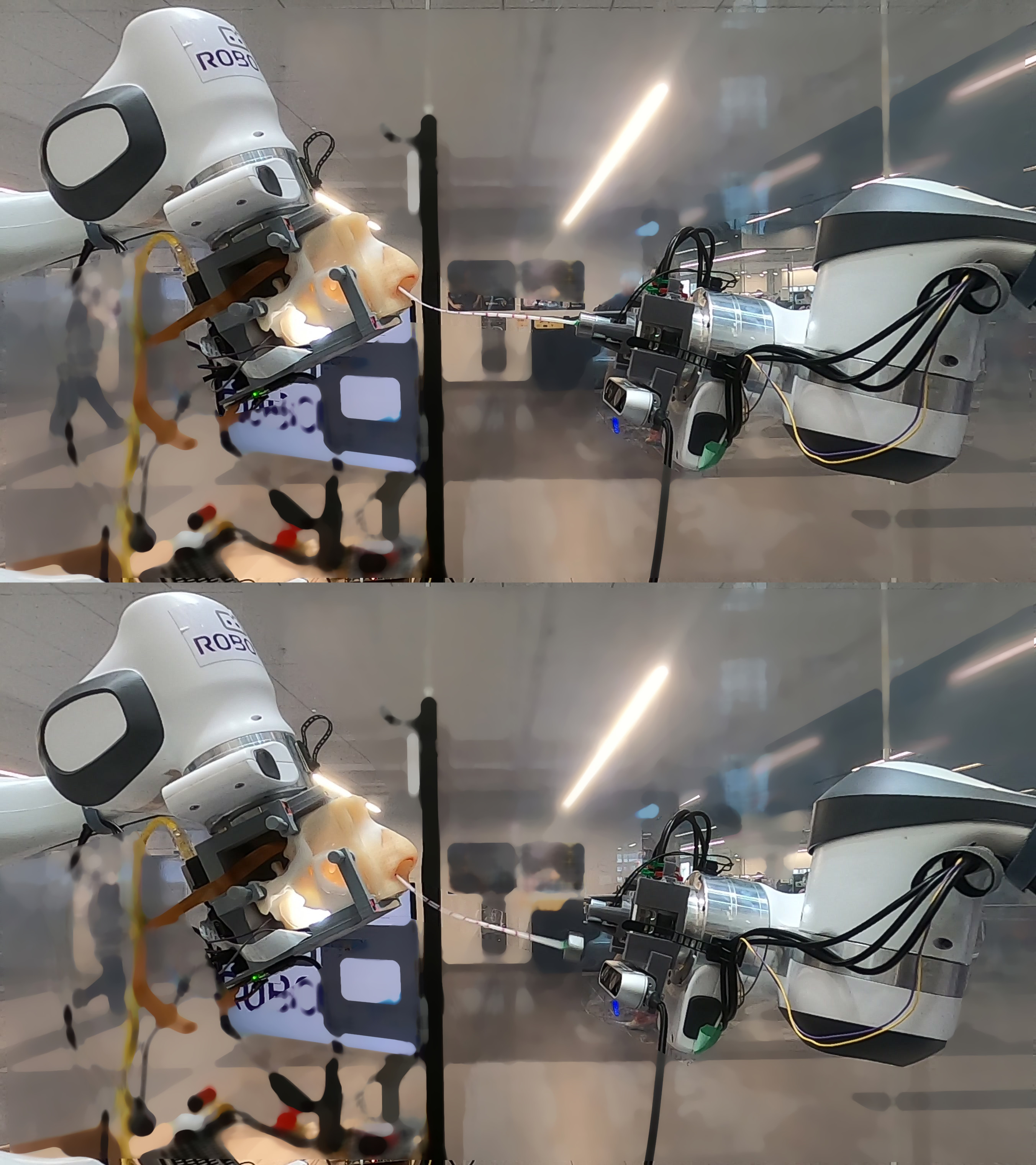}   \caption[Example of safety observer]{Example of the safety observer in use. Top: During the first few seconds of the contact stage, the swab tip becomes dislodged outside the nostril. Bottom: The observer detects a high impulse in force and aborts the procedure within 0.2 seconds by depowering the electromagnet and moving away.}
  \label{fig:int:abort}
\end{figure}

Our system also includes an observer to detect if any of two distinct conditions occur during the contact phase. If either of these safety conditions trigger during the insertion process, the electromagnet depowers, which releases the swab, and the system immediately transitions to stage 3, where the arm then retreats away from the face and returns to its starting position. 
The first condition detects if the swab becomes wedged on the nasal vestibule or outside the nostril. This could occur if the head moved suddenly after the visual servo component has completed. In this case, successful insertion is impossible, and it is better to abort the
insertion and try again. This scenario is detected with a simple rule: if there is a high spike in force measured early 
during the insertion, specifically if there is a force measurement above 0.5 N during the
first 40 mm of displacement. Fig. \ref{fig:int:abort} shows a sequence where this component triggered and ended the trial. The second condition monitors whether the total measured force exceeds a maximum
threshold of 2.5 N.

\section{Head motion simulator}
    \label{sec:head_model}

Interest in head motion is present in both social interaction and biomechanical fields, with models typically being targeted for a
particular task or action. For example, one area of active research is the generation of realistic facial poses and expressions for speaking
tasks \cite{Sadoughi2016-361} \cite{BenYoussef2013-362} \cite{Mittal2023-383} \cite{Chen2020-384}. There are also applications to mimicking or teleoperating heads in humanoid robots to
look more natural \cite{Agarwal2016-385} \cite{Mulavara2002-387} \cite{Antonelli2014-388}. In biomechanics, the head is but one extremity that is measured for tasks involving
motion of the body (e.g., \cite{Hirasaki1999-386} \cite{Mulavara2002-387}). In terms of kinematic modelling, the head will move along with the torso, but can also independently move by rotation about the neck. Generally, this rotation is modelled with a 3D spherical joint within the neck \cite{Sadoughi2016-361} \cite{Mulavara2002-387} \cite{Leva1996-350}.

Considering the absence of literature examining head motions of patients whom are being subjected
to the NP swab test, we look towards stochastic motion models for general head motion.
The study by Heuring and Murray~\cite{Heuring1999-363} is most relevant to our scenario. The authors proposed and evaluated two stochastic processes to model 6-DOF head motion during an image-based motion capture study: the Wiener process and
the Ornstein-Uhlenbeck process~\cite{Uhlenbeck1930-364}. Ultimately, the authors found that the Ornstein-Uhlenbeck process
autocorrelated much better measured head motion trajectories than the Wiener process, which we will base our head motion model on for our simulator.

\begin{figure}
  \includegraphics[width=\linewidth]{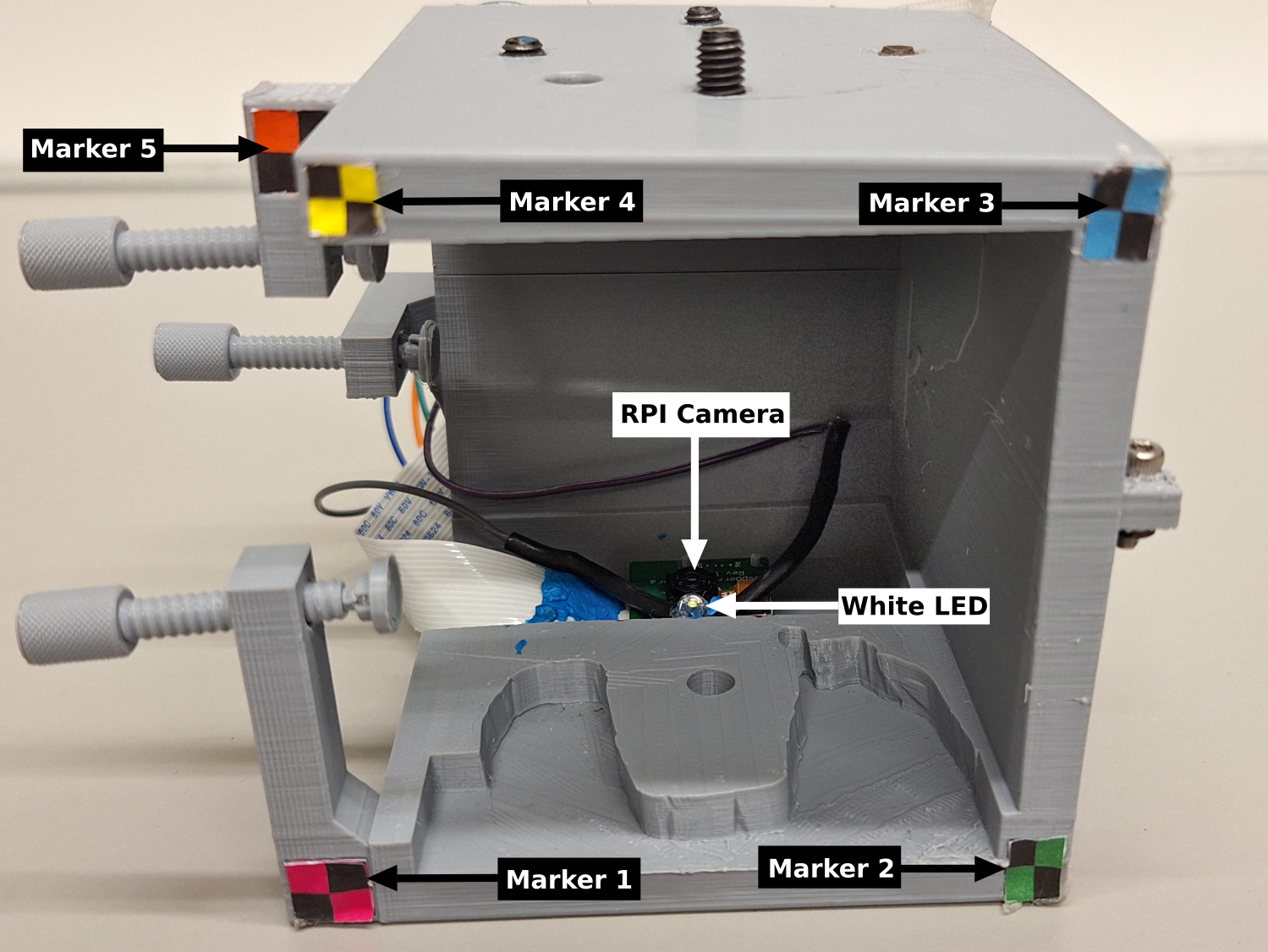}
  \caption[Phantom fixture for robot arm]{Front view of the empty fixture. The fixture contains five fiducial markers to help emulate the head pose estimation. A small Raspberry Pi camera is placed underneath the nasopharynx to determine if the swab successfully makes contact.}
  \label{fig:head_rig}
\end{figure}

The second arm dedicated towards simulating head motion has a 3D printed PLA fixture attached to the flange, which holds a nasal cavity phantom that is described in the following paragraph. The fixture itself, as shown in Fig \ref{fig:head_rig}, holds five fiducial markers to support pose estimation. At the
back of the phantom, underneath where the nasopharynx sits, a Raspberry Pi camera and an illuminating white LED take images of the
nasopharynx in order to observe whether the swab makes contact.  

The testing phantoms were based off of the designs by
Sanan{\`e}s \textit{et al.}~\cite{Sananes2020-91}. This phantom was designed to be a realistic training apparatus for training HCWs to take NP swab samples during the COVID-19 pandemic. The phantom is fabricated using a
Stratasys (Minnesota, USA; Rehovot, Israel) J750 polyjet printer
using Vero, a rigid material, and Agilus30, a rubber-like material to simulate the material properties of tissue. In
addition, we followed the recommendations of Sanan{\`e}s \textit{et al.}~\cite{Sananes2020-91} and applied petroleum jelly to the interior of the
cavity to simulate the lubricating effect of mucus. However, when comparing the phantom to 3D models
\cite{Keustermans2018-172} and CT scans \cite{Kuo2020-357} \cite{Smith2018-359} published in the literature, it appears that the inferior turbinate in the phantom is shrunken so the corresponding passage the swab would travel through is wider. The inventors may have
done this to facilitate easier training of practitioners, but we would like to evaluate our system with a narrower passage to ensure it is robust to different morphologies. As a result, we created a
second version of the phantom by coregistering the 3D model of the nasal cavity by
Keustermans \textit{et al.}~\cite{Keustermans2018-172} and then manually expanding the inferior turbinate using 3D modelling software Blender to
take up approximately the same volume. Throughout this manuscript we will designate the original phantom by Sanan{\`e}s \textit{et al.}~\cite{Sananes2020-91} as Phantom A and our modified one as Phantom B.

\begin{figure}
  \begin{subfigure}{0.49\linewidth}
    \includegraphics[width=\linewidth]{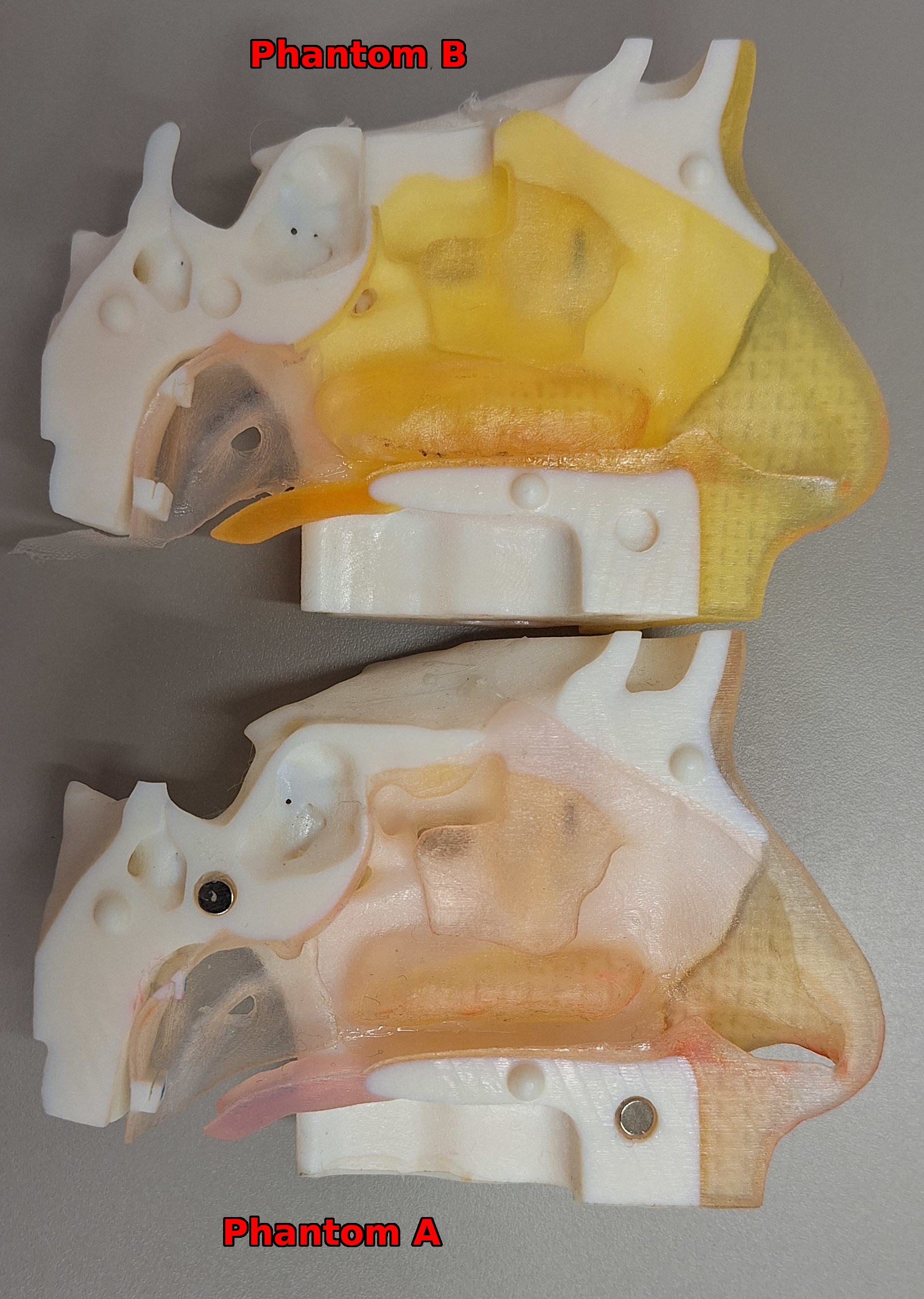}
    \caption{}
  \end{subfigure}
  \begin{subfigure}{0.49\linewidth}
  \includegraphics[width=\linewidth]{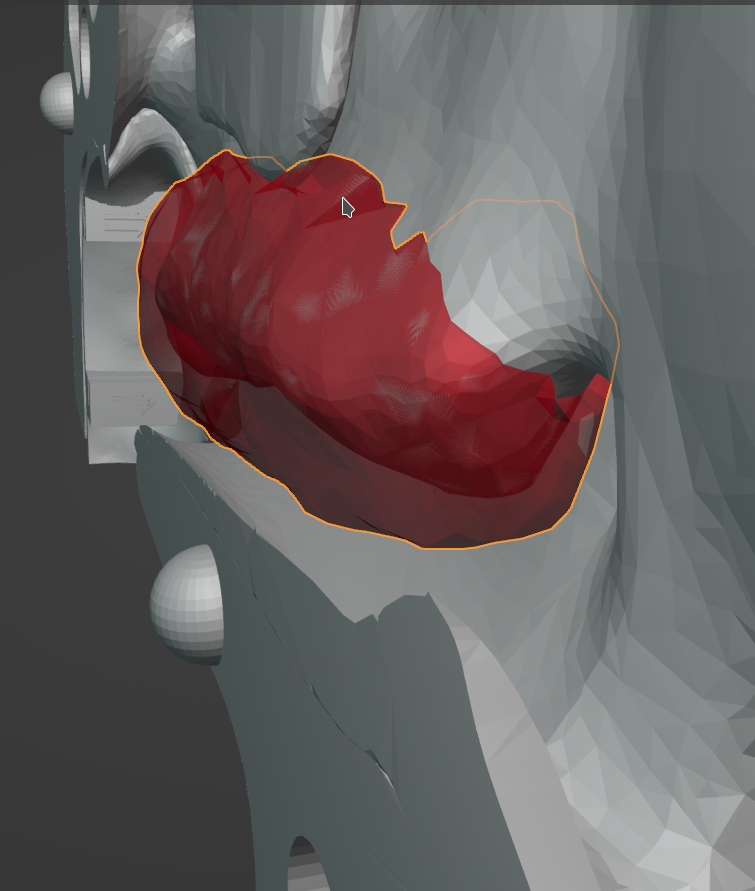}
    \caption{}
  \end{subfigure}
  \caption[Modifications to phantom]{a) Phantom A: the original phantom from Sanan{\`e}s \textit{et al.}~\cite{Sananes2020-91} and Phantom B: the modified version to
    have enlarged inferior turbinates, constricting the lower passage. 
    b) View of volume added (red) to Phantom B to more closely resemble the nasal cavity volume from Keustermans \textit{et al.}~\cite{Keustermans2018-172}.}
  \label{fig:phantom-figures}
\end{figure}

\begin{figure}
  \includegraphics[width=\linewidth]{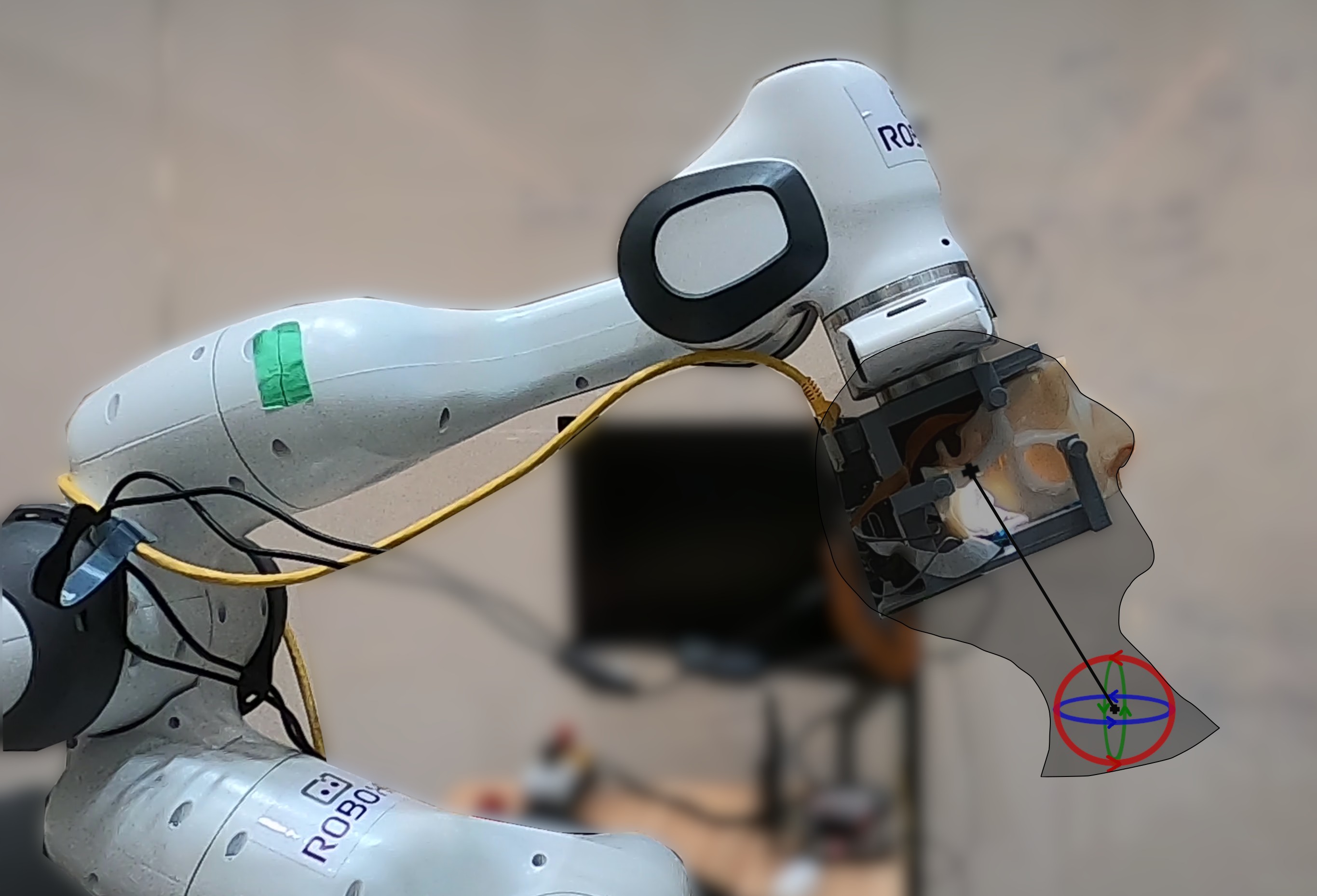} 
  \caption[Kinematics of head model]{Visualization of the kinematic model of the head, with a black silhouette of the head added for illustrative purposes. The arm moves the head as if it
    were constrained to a spherical neck joint 121.4 mm below the centre of the phantom.}
  \label{fig:head_kin}
\end{figure}

To emulate head motion, the arm will be moving with a certain set of kinematic constraints. Specifically, the arm will be rotating the centre of the phantom about 3D spherical neck joint located below the head, as shown in Fig.~\ref{fig:head_kin}.
The length of the virtual link connecting the centre of the head and the neck joint was set to 121.4 mm, the average between the centre of the mass of the head and the cervicale according to the study by de Leva~\cite{Leva1996-350}. While the neck is capable of 6 DOF \cite{Doss2023-433}, we adopt the simplified 3 DOF rotation model that is typically used to model the active range of motion for a standing person~\cite{Ferrario2002-434}.
A proportional joint velocity controller is used to move the head to the target rotation. Suppose that the
current angles of the virtual joint are $(\alpha,\beta,\gamma)$ and the target angles are $(\alpha_d,\beta_d,\gamma_d)$. The control law for the phantom arm is then defined as
\begin{equation}
  \dot{\mathbf{q}} = k\mathbf{J}^{\dagger}( \bm{\kappa}(\alpha_d,\beta_d,\gamma_d) - \bm{\kappa}(\alpha,\beta,\gamma)),
\end{equation}
where $\bm{\kappa}(\alpha,\beta,\gamma)$ is the forward kinematics function that computes the Cartesian pose
of the end-effector given the angles of the virtual neck joint, $\dot{\mathbf{q}}$ is the joint velocity vector, $\mathbf{J}$ is the robot Jacobian, and $k=0.5$ is the gain scalar.

\begin{figure}
  \includegraphics[width=\linewidth]{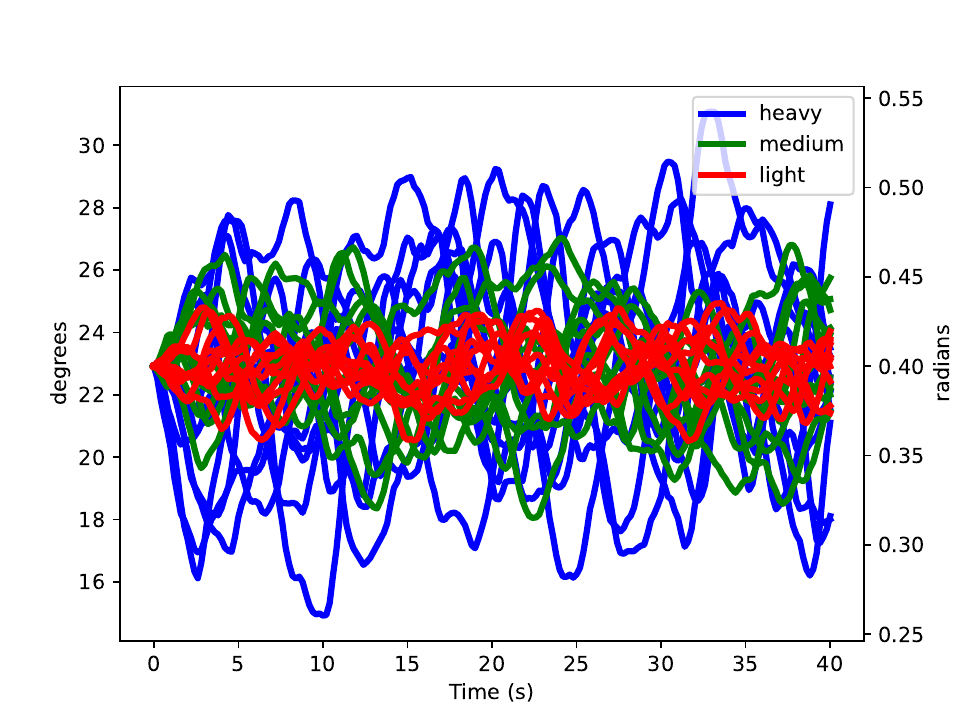}
  \caption[Simulated Ornstein-Uhlenbeck head motion]{Plots showing ten example motions for the proposed modified Ornstein-Uhlenbeck process
    for  \textbf{Light}, \textbf{Medium}, and \textbf{Heavy} head motion. The plots are
  shown for the angle $\alpha$ with a set point of 0.4 rad., but are similar for $\beta$ and $\gamma$ that would have a set point
of 0 rad. instead.}
  \label{fig:OU}
\end{figure}

The motion generation model is designed under the presumed scenario that the patient is cooperating and therefore attempting to hold their
head at a target pose, with some natural, random head motion. Based on the results from Heuring and Murray~\cite{Heuring1999-363}, we chose to model this natural head
motion by using a modified Ornstein-Uhlenbeck process, which is a stochastic dynamical system that
is generally used to simulate random processes, such as Brownian motion. Ultimately, the stochastic motion model we define for each angle is
\begin{equation}
  \begin{aligned}
    \dot{\theta} &= v_\theta \\
    \dot{v_\theta} &= -\mu(\theta-\bar{\theta}) -\psi v_\theta + \sqrt{2\psi}\mathcal{N}(0,\sigma_\theta),
\end{aligned}
\end{equation}
where $\theta \in (\alpha,\beta,\gamma)$, $\overline{\theta}$ is the setpoint angle with
($\overline{\alpha}=0.4$ rad, $\overline{\beta}$ = 0.0 rad, $\overline{\gamma}$ = 0.0 rad), $\mathcal{N}(0,\sigma_\theta)$ is a zero mean
random sample from a Gaussian Normal distribution with
standard deviation $\sigma_\theta$, $\psi$ is the velocity damping coefficient, and $\mu$ is the setpoint attraction coefficient. In the standard Ornstein-Uhlenbeck model, the change in velocity stochastically changes with the term
$\sqrt{2\psi}\mathcal{N}(0,\sigma_\theta)$ and is
damped by adding the term $-\psi v_\theta$. We modified this by adding an attractor term
$-\mu(\theta-\bar{\theta})$ so the head does not diverge from too far from the initial pose.
Thus, the model has three parameters that control the behaviour of the process: $\mu$, $\psi$, and $\sigma_\theta$. We decided on
three sets of parameters that would correspond to \textbf{Light} head motion ($\sigma_\theta=0.5$,
$\psi=1.0$, $\mu = 1.0$), \textbf{Medium}
head motion ($\sigma_\theta = 0.7$, $\psi = 1.0$, $\mu = 0.5$), and \textbf{Heavy} head motion ($\sigma_\theta = 1.2$
$\psi = 0.75$, $\mu = 0.5$). Fig. \ref{fig:OU} shows a comparison of the simulated angles over
time for the selected levels of head motion.

\section{Experiments}
\label{sec:int_experiments}

In our experiments there are a total of four categories of independent variables that were combinatorially varied to evaluate the controller over different conditions. These variables are: \begin{enumerate}
\item Controller configurations: \{\textit{S1.0}, \textit{S1.5}, \textit{S2.0}, \textit{D1.0}, \textit{D1.5}, \textit{D2.0}\} (see Section \ref{sec:compliant_controller}).
\item  Degree of head motion: \{\textbf{Light}, \textbf{Medium}, \textbf{Heavy}\} (see Section \ref{sec:head_model}).  One limited, non-repeated trial was also run with no movement with no head motion (\textbf{None}) for comparison purposes.
\item Phantom model: \{A, B\}(see Section \ref{sec:head_model}) .
\item Nostril side: \{Left, Right\}.
  \end{enumerate}
In total, there were 564 insertion trials over these configurations. For each combination of the first three variables, 10 trials
were done for the left side
of the nostril, and 5 were done for the right side. 

\begin{figure*}
  \includegraphics[width=\linewidth]{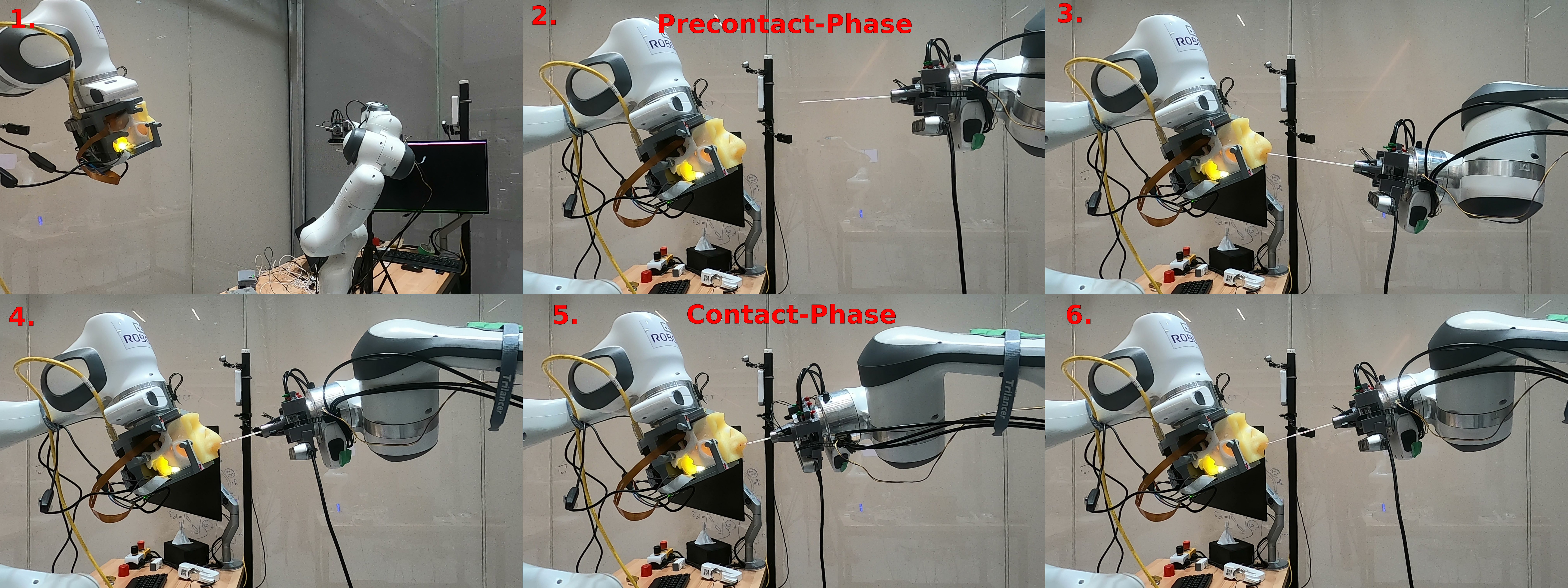}
\caption[Sequence of motion during a trial]{Illustration of
    the entirety of the integrated test. Frames 1-3: the pre-contact phase starts from the
    sentry mode and ends with the swab tip about to enter the nostril. Frames 4-6: the contact-phase
    first inserts the swab into the nasal cavity until it reaches the nasopharynx. After rotating to
    collect a sample, it exits the nasal cavity.}
  \label{fig:int:sequence}
\end{figure*}

\begin{figure*}
    \includegraphics[width=\linewidth]{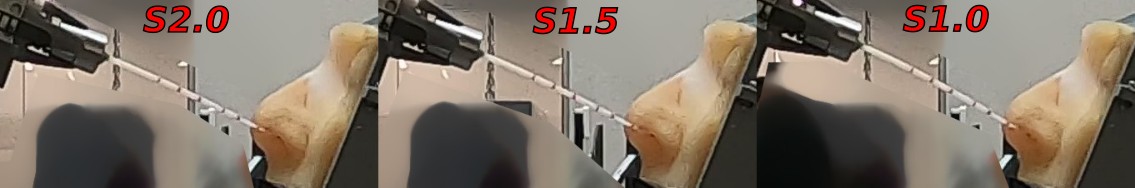}
    \caption[Trial comparison of gain levels]{Frames taken
    during a trial with different controller configurations. Notice how the lower gain controllers result in the swab having a greater bend, and the upper shafts are more misaligned from the nostril compared to the higher gain ones.}
  \label{fig:int:gain_compare}
\end{figure*}

\begin{figure}\begin{center}
    \includegraphics[width=0.75\linewidth]{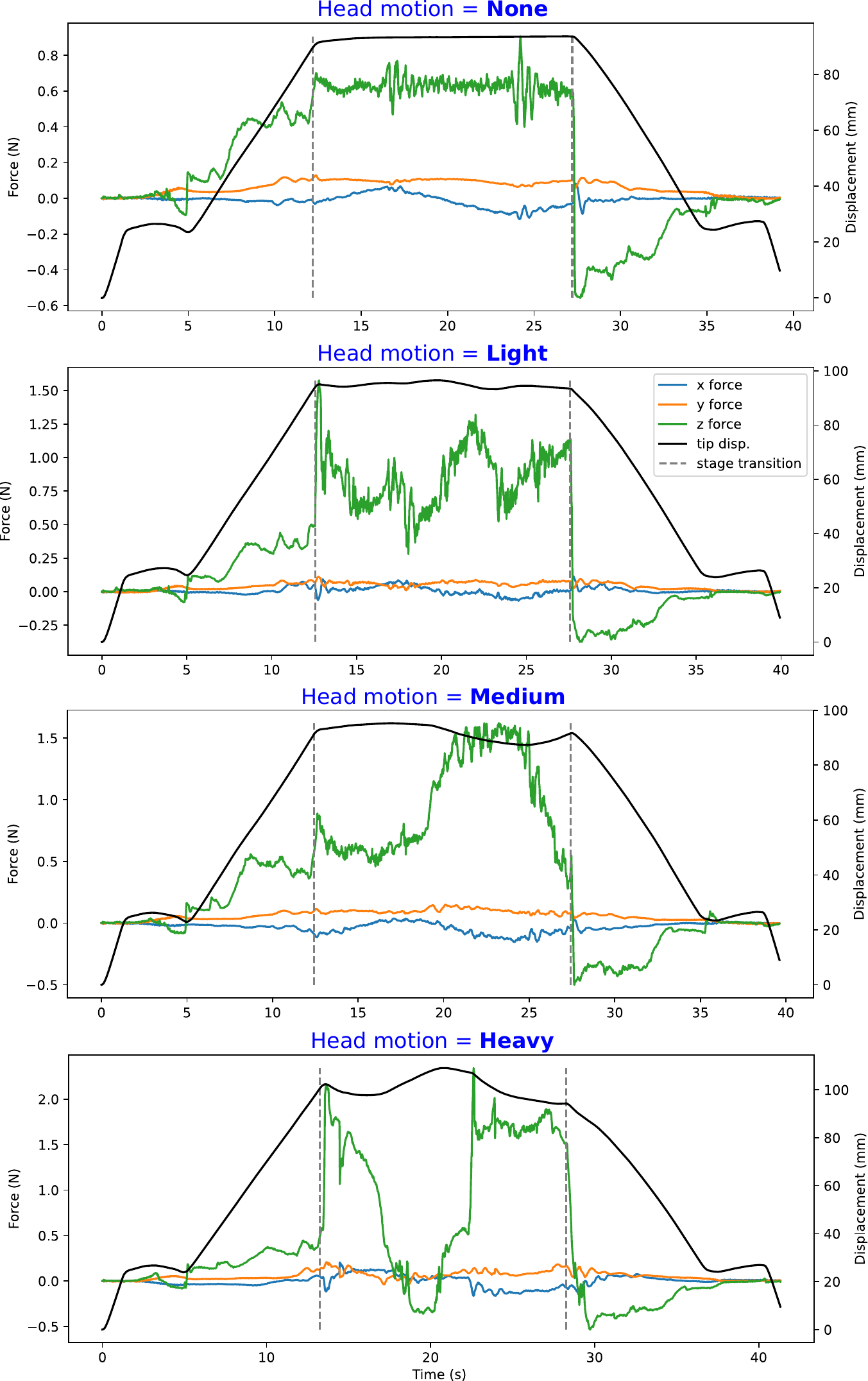}
    \end{center}
  \caption[Graphs comparing forces and tip displacement with head motion]{Graphs comparing the measured forces and the displacement of the tip for different levels
  of head motion for controller \textit{D2.0} with Phantom B on the left side of the
  nostril. Generally higher amounts of head motion will result in the swab moving more to compensate
for the forces of the rocking head. The three stages are divided according to the vertical dashed lines.}
  \label{fig:graph_head_compare}
\end{figure}

The experiment trials were conducted during February 2024 at the University of Waterloo RoboHub.
The physical setup involved both of the Franka Emika arms, with the table holding the phantom wielding arm being placed into the workspace of the swab wielding arm. At the
beginning of each trial, the swab arm would begin in its sentry position and then the entire integrated procedure would execute (i.e., the
visual servo pipeline followed by the compliant control pipeline). The computers controlling the head robot and the swab robot were networked so that
the latter could automatically trigger head motion generation at the end of the sentry stage, after the swab has been segmented and calibrated. Fig. \ref{fig:int:sequence}
shows a collection of images taken during the trials. Fig. \ref{fig:int:gain_compare} compares a few different controller configurations that also alludes to the transverse gain
has an impact on trial outcomes. The total 
displacement required to reach the nasopharynx for a non-moving head 
was 93.1 mm for the left nostril and 94.2 mm for the right nostril.
In terms of the total duration of the test, the pre-contact phase required varying amounts of time for the visual servo to converge, depending on the level of head motion. The average times, after the sentry stage, were as follows: \textbf{None}: 11.3~s, \textbf{Light}: 11.9~s, \textbf{Medium}: 13.7~s, and \textbf{Heavy}: 18.6~s. The contact phase, on average, took a total of 39.5~s.
Fig. \ref{fig:graph_head_compare} plots the forces and displacement of the swab tip and compares how
these quantities fluctuate with different head motion configurations.
The independent variables clearly have an impact on the results of the trial, which we will
quantify via the measures defined below.

\subsection{Measures}
\label{sec:metrics}
The evaluated measures for the trials are largely divided into the two stages of insertion. For each of the measures,
we will be examining the statistical significance of the independent variables using four-way ANOVA testing for
the continuous measures and chi-square test for the discrete measure. Consequently, we report the F statistics for ANOVA, $\chi^2$ statistic for the chi-square test, and both of their associated p-values. The critical p-value for these
tests was adjusted with Bonferroni correction~\cite{Dunn1961-382} based on the number of tests per measure:
$0.05/4 = 0.0125$.

The measures corresponding to stage 1 (insertion) are:
\begin{enumerate}[label=\roman*)]
\item Transverse force ($\Downarrow$)
\item Axial force ($\Downarrow$)
\item Time to reach nasopharynx ($\Downarrow$)
\item Whether the nasopharynx was reached ($\Uparrow$)
\end{enumerate}
The $\Downarrow$ means that the lower values for that measure are better, while $\Uparrow$ indicates that higher values are better.
The first two measures are computed by taking the average forces applied in the transverse (x and y) and axial (z) directions. These values are interesting to monitor because they reflect the pressure that the swab is exerting within the nasal
cavity. The third measure, the time to reach the nasopharynx, shows whether the controller slows down or becomes stuck during stage 1,
which would be an uncomfortable situation for a human patient.
The fourth measure is obviously important because the ultimate objective is for the swab tip to reach the nasopharynx. If it does not, then
the sample collection will be suboptimal.

\begin{figure}
  \begin{center}
    \includegraphics[width=\linewidth]{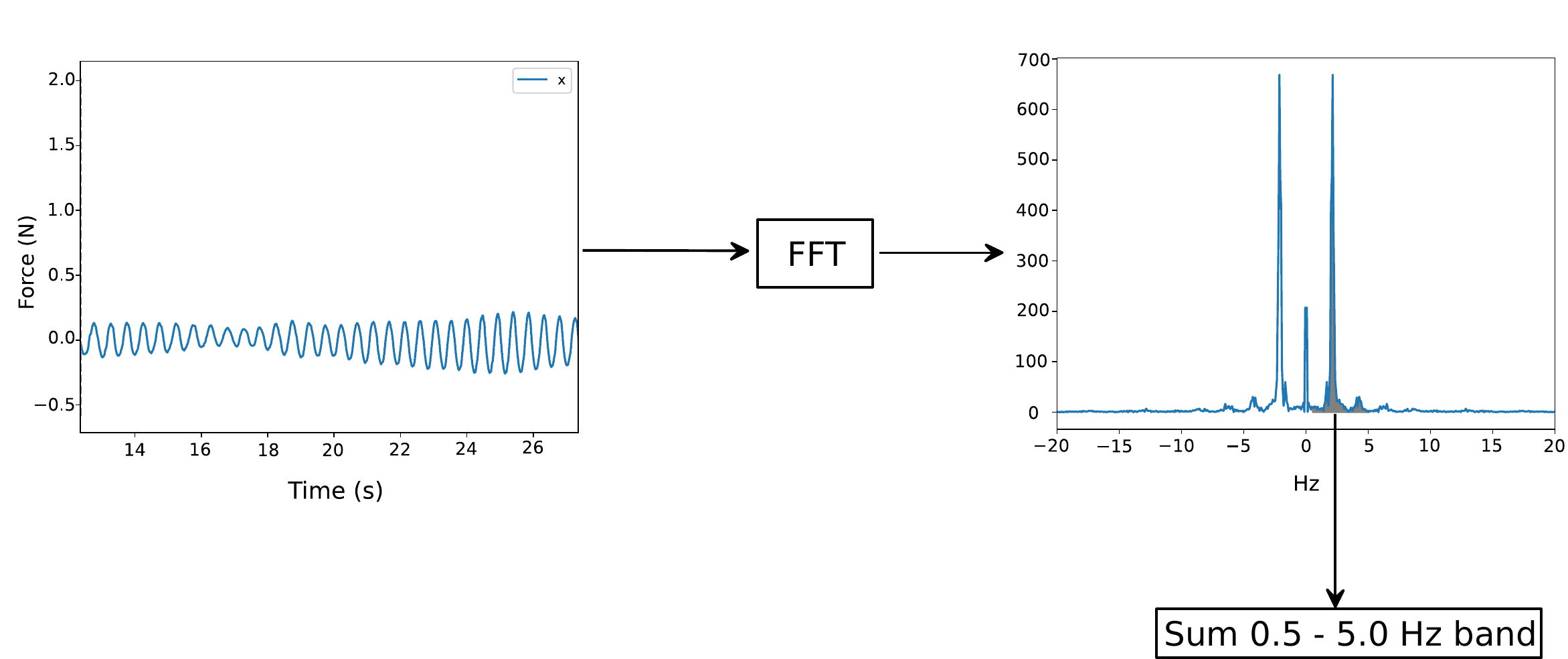}
    \end{center}
  \caption[Illustration of computation of oscillation coefficient]{An illustration of how the oscillation coefficient is computed. The left image shows the x-axial force profile for a trial with
    high oscillation. In the frequency domain, this is characterized by a large spike around 2 Hz. The measure is computed by summing up the
    bins between the 0.5 to 5 Hz range.}
  \label{fig:oscil_example}
\end{figure}

The measures corresponding to stage 2 (collection) are:
\begin{enumerate}[resume,label=\roman*)]
\item Transverse force ($\Downarrow$)
\item Axial force ($\Downarrow$)
\item Oscillation coefficient ($\Downarrow$)
\end{enumerate}
The fifth and sixth measures are computed in the same way as the first and second, except they are averaged over the
duration of stage 2. In configurations where the force gain was poorly tuned, we noticed that the forces measured on the transverse axes would
oscillate during trials, which would cause the end-effector to visibly shake. We designed the oscillation coefficient as a way of measuring
and comparing this phenomenon between different configurations.
The first step of computing the oscillation coefficient is taking the discrete
Fourier transform of the x-axis force over the duration of stage 2. We observed that the high
oscillation trials could be characterized by larger spikes in the frequency domain, as shown in
Fig. \ref{fig:oscil_example}. As a result, our final measure is computed by summing the
frequency bins between 0.5 Hz to 5 Hz, the band where the oscillations typically occupied.

In the remainder of this section we summarize how each of the measures are impacted by the independent variables. Tables \ref{tab:stage1_metrics} and \ref{tab:stage2_metrics} show the results from the aforementioned statistical tests and the summary statistics of the measures partitioned by each variable. Figures \ref{fig:s1_1}, \ref{fig:s1_2}, \ref{fig:s1_3}, \ref{fig:s2_5}, \ref{fig:s2_6}, and \ref{fig:s2_7} are categorical plots that partition the results of the trials according to the independent variables, which are formatted according to footnote\footnote{The 2 $\times$ 3 Violin plots show the distribution of the measures and are partitioned according to the four groups of independent variables. The rows denote the side of the phantom and the columns denote the level of head motion. The x-axis denotes the controller type and the colour of the plot denotes the phantom used. Measures were not computed for trials that got stuck and did not reach stage 2.\label{ft:violin_expl}}.

\subsection{Stage 1: Measure i) Transverse force}
\label{sec:res_i}

\begin{figure}
      \includegraphics[width=\linewidth]{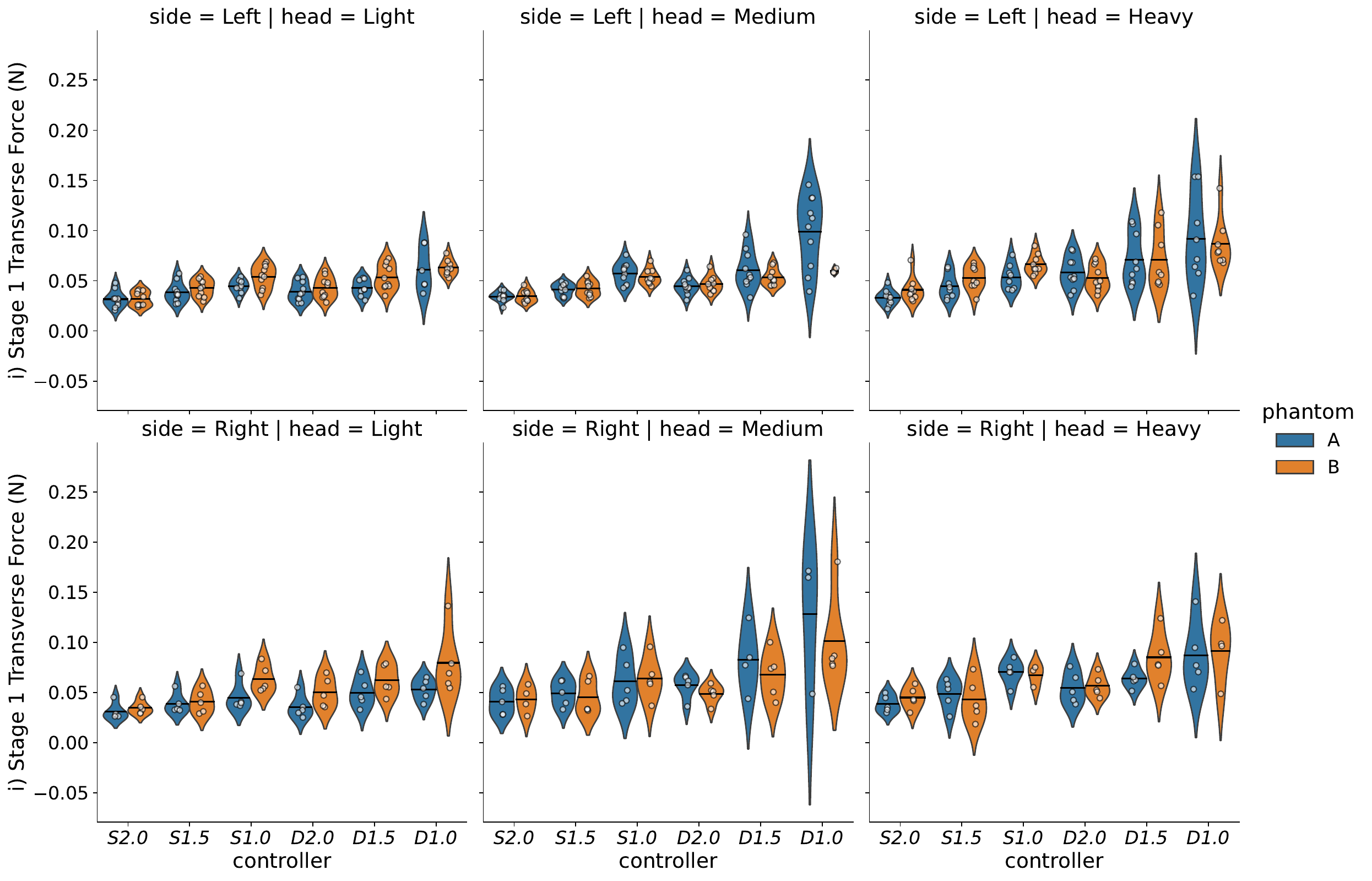}
      \caption[Measure i) violin plot]{Measure i) Violin plots showing distribution of transverse forces during stage 1. See footnote \textsuperscript{\ref{ft:violin_expl}} in Section \ref{sec:metrics} for formatting.}       
      \label{fig:s1_1}
    \end{figure}

For the stage 1 transverse force, we observe that there are significant differences among the control, head motion, and
side variables, but the difference between the two phantoms is insignificant, as
summarized in Table \ref{tab:stage1_metrics} and Fig. \ref{fig:s1_1}. Controllers with lower transverse force gain have increased
in average transverse force because the controller will favour the target waypoint position at the
cost of higher force. It also appears that the spread of forces raises for the lower gain controllers as well. Heavier head motion seemed to
lead to disturbances that resulted in higher transverse force as well. The forces from 
left side nasal cavity passage also appear to be slightly higher than the right side.

\subsection{Stage 1: Measure ii) Axial force}

\begin{figure}
      \includegraphics[width=\linewidth]{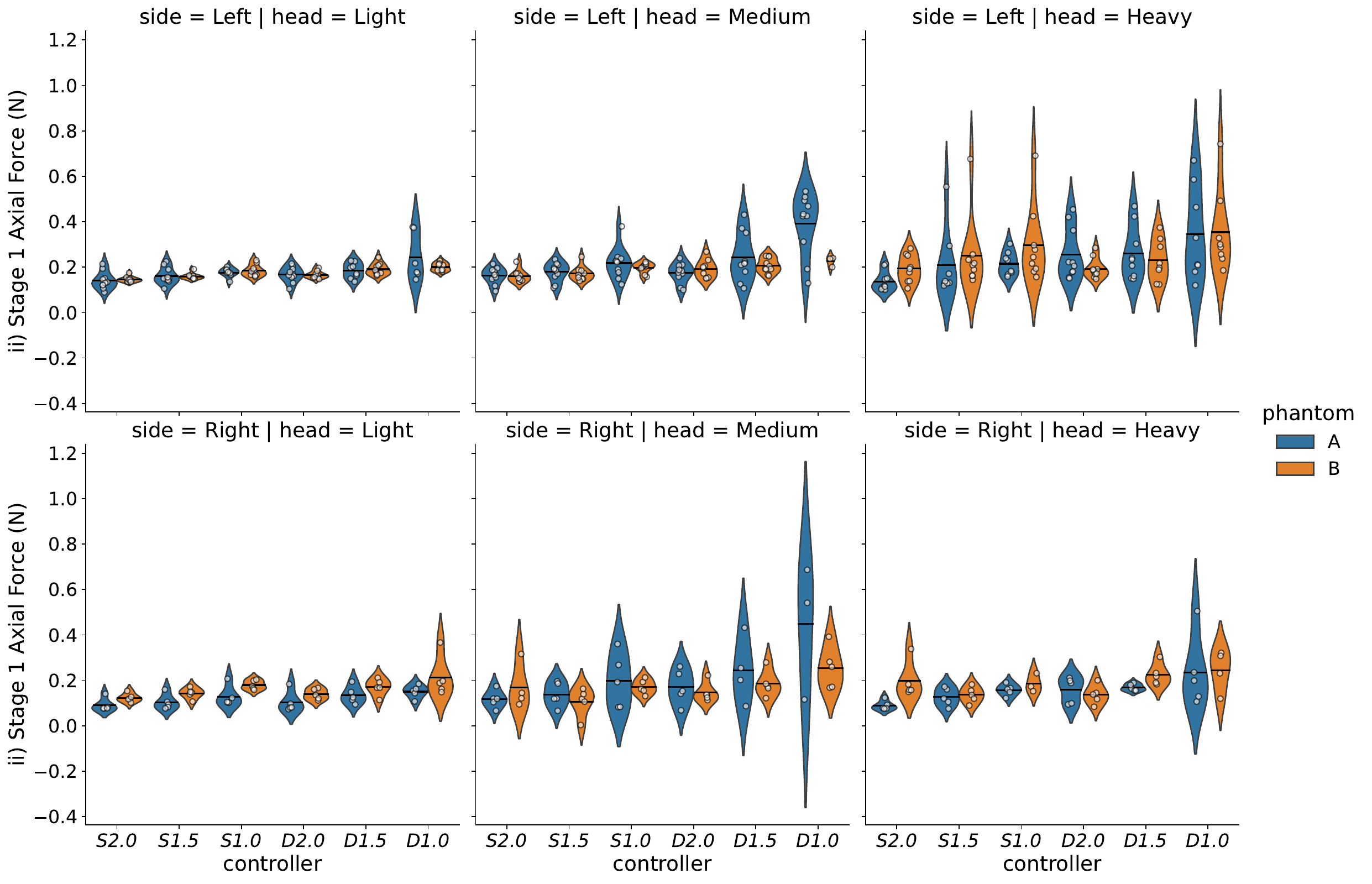}
      \caption[Measure ii) violin plot]{Measure ii) Violin plots showing distribution of axial forces during stage 1. See footnote \textsuperscript{\ref{ft:violin_expl}} in Section \ref{sec:metrics} for formatting.}
      \label{fig:s1_2}
    \end{figure}

Table \ref{tab:stage1_metrics} and Fig. \ref{fig:s1_2} shows the axial force during stage 1.  The control, head, and side variables
were significant, while the phantom variable was not. The outcomes appear to be correlated with those
to those for the transverse force. As the axial force gain was the same for all controllers, this
appears to indicate that the axial force transmission from Coulomb friction is a visible effect
during stage 1 as the swab moves forward through the nasal cavity.

\subsection{Stage 1: Measure iii) Time to reach nasopharynx}
\label{sec:res_iii}

\begin{figure}
      \includegraphics[width=\linewidth]{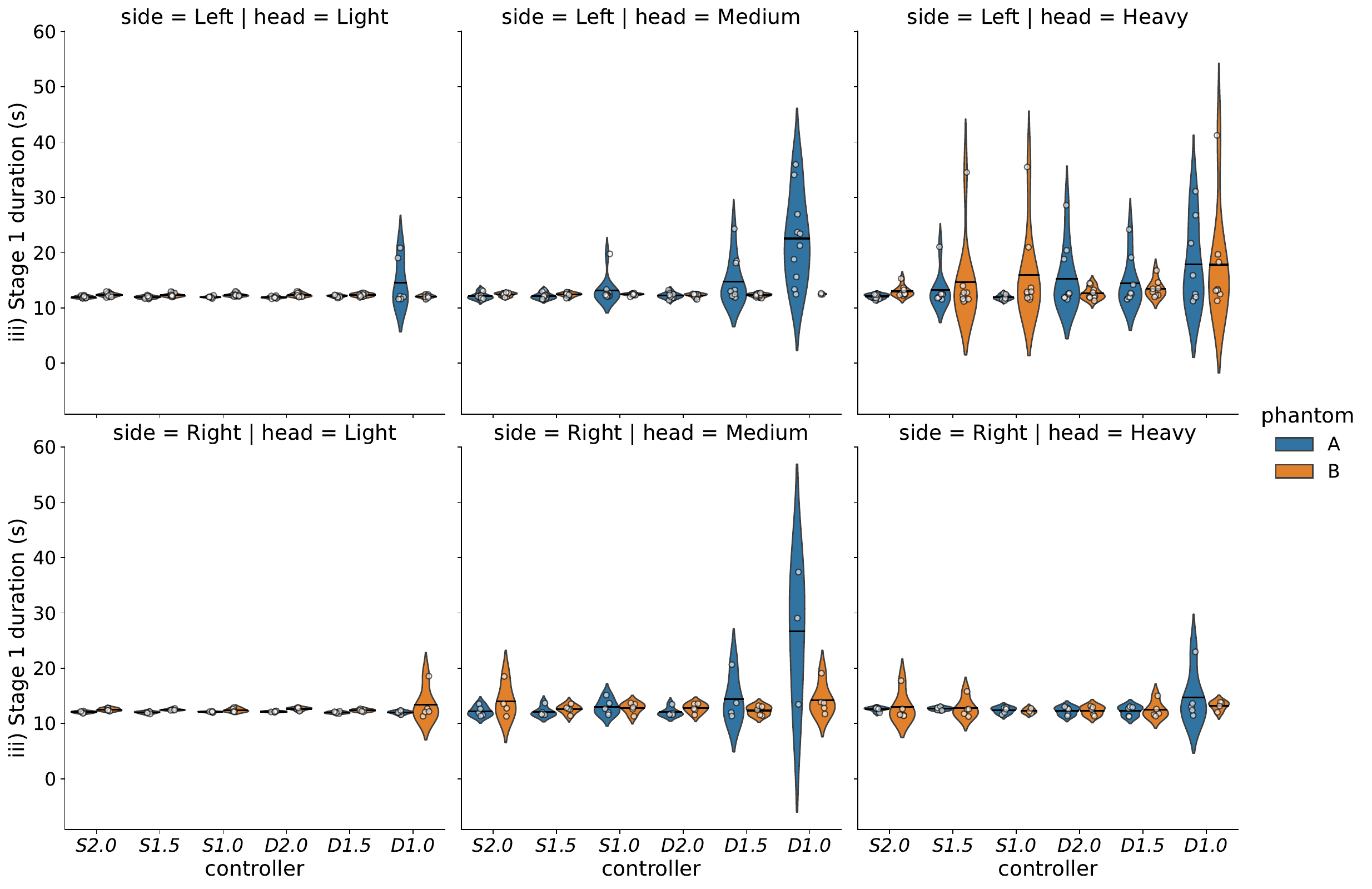}
      \caption[Measure iii) violin plot]{Measure iii) Violin plots times required to reach NP. See footnote \textsuperscript{\ref{ft:violin_expl}} in Section \ref{sec:metrics} for formatting.}
      \label{fig:s1_3}
    \end{figure}

Table \ref{tab:stage1_metrics} and Fig. \ref{fig:s1_3} shows that the time to reach the nasopharynx, which was largely the same for all the controllers, except for
\textit{D1.0}. For \textit{D1.0}, it appears that the axial forces were high enough to cause a
significant slowdown via the mechanism in (\ref{eq:slowdown_zaxis}), although this effect appears to be quite variable according to the high
standard deviation. It also that heavier head motions could also result in the increased time to reach the nasopharynx.

\subsection{Stage 1: Measure iv) Reached nasopharynx}
\label{sec:res_iv}

\begin{table}
  \caption[Breakdown of failures to reach stage 2 by controller configuration]{Breakdown of failures to reach stage 2 by controller configuration. Categories of failures include those that became stuck (did not progress during insertion) due to (\ref{eq:slowdown_zaxis}) reaching zero, one case where the procedure aborted due to the force threshold being exceeded, and those where the stage 2 observer triggered prematurely (before the swab tip reached the nasopharynx).}
  \label{tab:siv_breakdown}
  \begin{center}
\begin{tabular}{|c|c|c|c|}\hline
Controller & Stuck & Overload & Premature\\\hline
  \textit{S2.0} & 1 & 1 & 0\\
  \textit{S1.5} & 2 & 0 & 1\\
  \textit{S1.0} & 5 & 0 & 2\\
  \textit{D2.0} & 1 & 0 & 3\\
  \textit{D1.5} & 6 & 0 & 11\\
  \textit{D1.0} & 19 & 0 & 34\\\hline
\end{tabular}
\end{center}
\end{table}

Table \ref{tab:stage1_metrics} shows the proportion of trials that reached the nasopharynx according to each variable.
Statistically, only the controller configuration had a significant impact on the success rate.
The lower gain controllers clearly
struggled with making it to the nasopharynx properly. For the dynamic controllers, we see a steady
increase in success rate from \textit{D1.0} to \textit{D2.0} as the gains also increase, and we can also see the
success rate for \textit{S1.0} being lower than the other static configuration. Despite not reaching the level of statistical significance, we do see a trend where lower head motion had better rates of success. Also, the left side had
about an 8\% lower success rate, but is still under the level of statistical significance.

It is important to analyze the specific events that could prevent the swab from reaching the nasopharynx. These obstacles can be categorized into three main types:
\begin{enumerate}[label=\alph*)]
  \item Controller stagnation due to axial force saturation
  \item Procedure abortion before reaching the nasopharynx due to exceeding the maximum safety force limit
  \item Premature triggering of the stage 2 observer before the swab actually reaches the nasopharynx
  \end{enumerate}
  Table \ref{tab:stage1_metrics} presents a breakdown of these events for each controller. For the low-gain controller, the most frequent issue was the premature triggering of the stage 2 observer, primarily caused by excessive axial force. Additionally, several insertions were terminated early when it became evident that the controller was stuck and not progressing. Only one instance occurred where the force threshold was exceeded before reaching the nasopharynx, although this did happen seven other times after stage 2 had begun.

  \begin{table*}
    \caption{Statistics and summary of stage 1 measures.}
    \label{tab:stage1_metrics}
    \begin{center}
      \begin{tabular}{|c||c|c||c|c||c|c||c|c|}\hline
  & \multicolumn{6}{|c||}{\textbf{Four way ANOVA} } & \multicolumn{2}{|c|}{ $\chi^2$ \textbf{test} } \\\hline
  & \multicolumn{2}{|c||}{\textbf{i) Transverse force}} & \multicolumn{2}{|c||}{\textbf{ii) Axial force}} & \multicolumn{2}{|c||}{\textbf{iii) Time to NP}} &  \multicolumn{2}{|c|}{\textbf{iv) Reached NP}} \\\hline
  \underline{Variable} & \underline{F} & $\underline{p}$ & 
  \underline{F} & $\underline{p}$ &  \underline{F} & $\underline{p}$ &  $\chi^2$ & $p$ \\
  Control & 70.5  & $5.1 \times 10^{ -56 }$ * & 28.3  & $2.5 \times 10^{ -25 }$ *
  &  14.1  & $6.1 \times 10^{ -13 }$ * & 165  & $9.2 \times 10^{ -34 }$ *\\
  Head motion & 20.9 & $9.1 \times 10^{ -13 }$ * & 14.2 & $6.7 \times 10^{ -9 }$ * & 6.99 & $1.3 \times 10^{ -4 }$ * & 8.09 & $4.4 \times 10^{ -2 }$ \\
  Side & 11.9 & $6.1 \times 10^{ -4 }$ * & 35.3 & $5.3 \times 10^{ -9 }$ * &  2.57 & $1.1 \times 10^{ -1 }$  & 6.2 & $1.3 \times 10^{ -2 }$ \\
  Phantom & 1.39 & $2.4 \times 10^{ -1 }$  & 0.0188 & $8.9 \times 10^{ -1 }$  & 2.46 & $1.2 \times 10^{ -1 }$  & 0.217 & $6.4 \times 10^{ -1 }$ \\\hline
  \multicolumn{9}{|c|}{\textbf{Summary stats} } \\\hline
  & \multicolumn{2}{|c||}{\textbf{i) Transverse force} (N)} & \multicolumn{2}{|c||}{\textbf{ii) Axial force} (N)} & \multicolumn{2}{|c||}{\textbf{iii) Time to NP} (s)} & \multicolumn{2}{|c|}{\textbf{iv) Reached NP}} \\\hline
  \underline{Control} & \underline{Mean} & \underline{Stdv.} &  \underline{Mean} & \underline{Stdv.} &  \underline{Mean} & \underline{Stdv.} & \multicolumn{2}{|c|}{\underline{Success rate} }\\
  \textit{S2.0} &   0.036   &   0.009 &    0.147   &   0.051 &    12.3   &   1.6&\multicolumn{2}{|c|}{0.979} \\
  \textit{S1.5} &   0.044   &   0.011 &    0.164   &   0.081 &    12.3   &   3.1&\multicolumn{2}{|c|}{0.968} \\
  \textit{S1.0} &    0.057   &   0.014 &     0.197   &   0.078 &     12.1   &   4.0 & \multicolumn{2}{|c|}{0.926}\\
  \textit{D2.0} &   0.048    &  0.013 &    0.174    &  0.063 &    12.5    &  2.4 &\multicolumn{2}{|c|}{0.957} \\
  \textit{D1.5} &   0.062    &  0.021 &    0.207    &  0.075 &    12.1    &  3.9 & \multicolumn{2}{|c|}{0.819} \\
  \textit{D1.0} &   0.082   &   0.035 &   0.278   &   0.149 &    12.8   &   8.9 & \multicolumn{2}{|c|}{0.426} \\\hline
  \underline{Head motion} & \underline{Mean} & \underline{Stdv.} & 
  \underline{Mean} & \underline{Stdv.} &  \underline{Mean} & \underline{Stdv.} & \multicolumn{2}{|c|}{ \underline{Success rate} }\\
  \textbf{None} & 0.047  &     0.012 & 0.163  &     0.046 &  11.7  &     2.5 &\multicolumn{2}{|c|}{0.875} \\
  \textbf{Light} & 0.046  &    0.016  &  0.162  &    0.047  & 11.6  &    3.0 & \multicolumn{2}{|c|}{0.906} \\
  \textbf{Medium} & 0.056 & 0.027  &  0.201 & 0.099  & 12.7 & 5.2     &      \multicolumn{2}{|c|}{0.822} \\
  \textbf{Heavy} &  0.060  &  0.025 &  0.216  &  0.120 &  12.9  &  5.4 & \multicolumn{2}{|c|}{0.806} \\\hline
  \underline{Side}& \underline{Mean} & \underline{Stdv.} & \
  \underline{Mean} & \underline{Stdv.} &  \underline{Mean} & \underline{Stdv.} & \multicolumn{2}{|c|}{ \underline{Success rate} }\\
  Left & 0.052  & 0.021 &  0.207  & 0.097 &  12.2  & 5.2  &    \multicolumn{2}{|c|}{0.817}  \\
  Right & 0.057  & 0.026 & 0.163  & 0.084 &  12.7  & 3.1 & \multicolumn{2}{|c|}{0.901} \\\hline
  \underline{Phantom} & \underline{Mean} & \underline{Stdv.} & 
  \underline{Mean} & \underline{Stdv.} &  \underline{Mean} & \underline{Stdv.} & \multicolumn{2}{|c|}{ \underline{Success rate} }\\
  A &  0.053 &   0.025  &  0.191 &   0.108  &   12.6 &   5.0 & \multicolumn{2}{|c|}{0.855} \\
  B &  0.055 &   0.021  &  0.192 &   0.080  &  12.1 &   4.2 & \multicolumn{2}{|c|}{0.837} \\\hline
\end{tabular}

    \end{center}
  \end{table*}

\subsection{Stage 2: Measure v) Transverse force}
\label{sec:s2_5}

\begin{figure}
  \includegraphics[width=\linewidth]{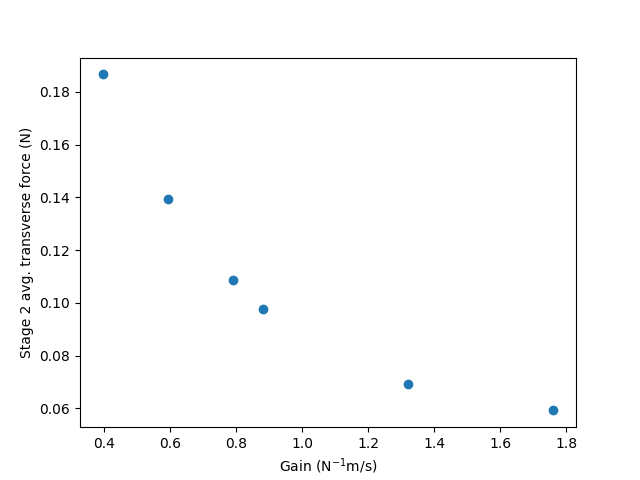}
  \caption[Average transverse force according to gain]{Average amounts of transverse force, with controllers ordered by their transverse gains at
    the nasopharynx.}
  \label{fig:transverse}
\end{figure}

\begin{figure}
      \includegraphics[width=\linewidth]{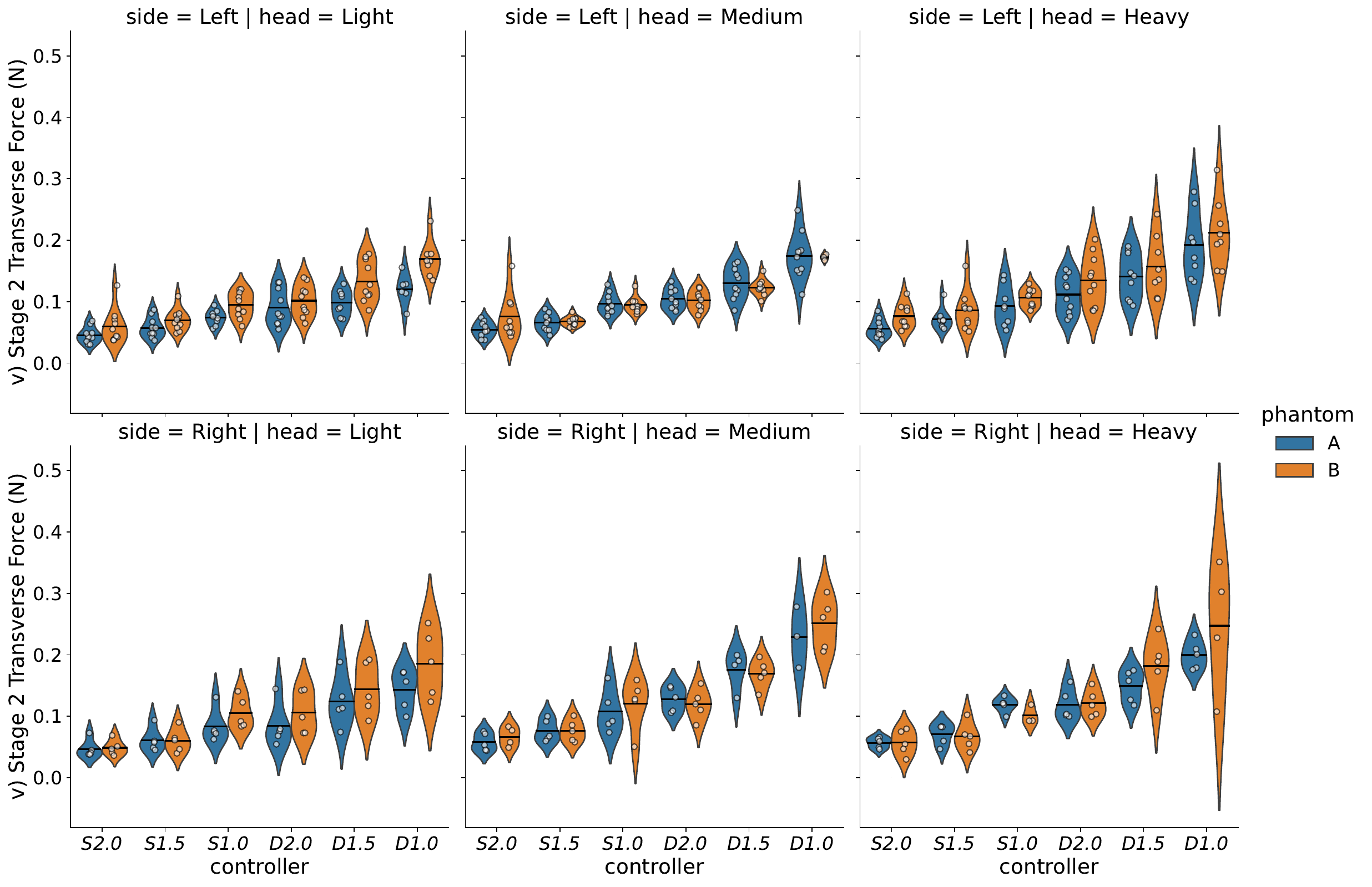}
      \caption[Measure v) violin plot]{Measure v) Violin plots showing distribution of transverse forces during stage 2. See footnote \textsuperscript{\ref{ft:violin_expl}} in Section \ref{sec:metrics} for formatting.}
      \label{fig:s2_5}
    \end{figure}

As shown in Table \ref{tab:stage2_metrics} and Fig. \ref{fig:s2_5}, all the variables have a significant impact on the average
transverse forces. As previously observed, the predominant factor appears to be the controller type, with a clear negative correlation between gain and transverse force. At stage 2, the
dynamic gain controllers have all reached their minimum gains at the nasopharynx, which are below
all the static gain controllers. We can observe a monotonic trend
between the six controller in Fig. \ref{fig:transverse}. In addition, we also see that heavier head motion moderately
raises both the mean and variance of the transverse forces. The left side of the nasal
cavity and Phantom B had somewhat greater transverse forces as well.

\subsection{Stage 2: Measure vi) Axial force}
\label{sec:s2_6}

\begin{figure}
      \includegraphics[width=\linewidth]{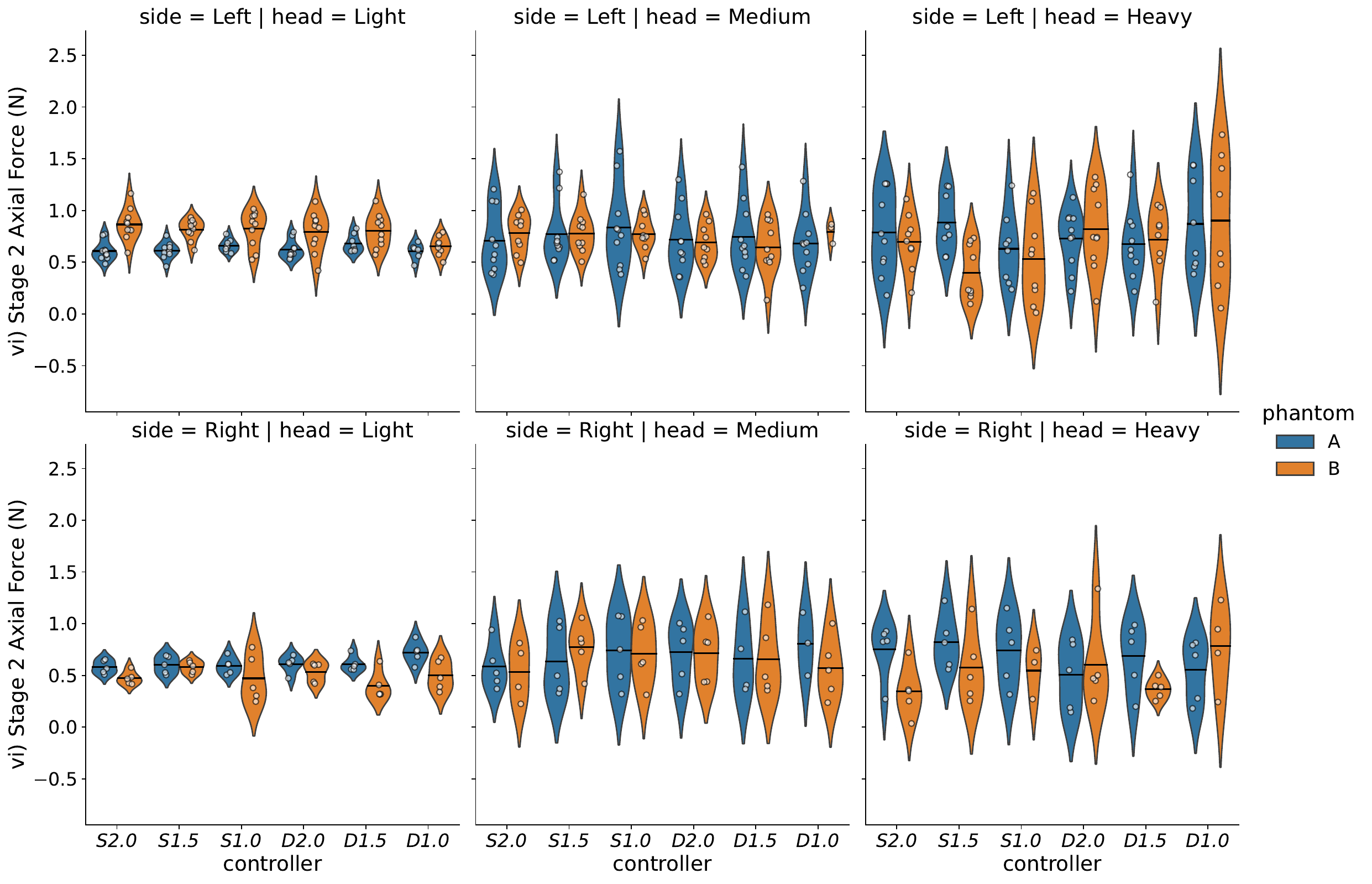}
      \caption[Measure vi) violin plot]{Measure vi) Violin plots showing distribution of axial forces during stage 2. See footnote \textsuperscript{\ref{ft:violin_expl}} in Section \ref{sec:metrics} for formatting.}
      \label{fig:s2_6}
    \end{figure}

From Table \ref{tab:stage2_metrics} and Fig. \ref{fig:s2_6}, it appears that the controllers have little
impact on the mean value for the axial forces. This contrasts the result from stage 1 (Measures i \&
ii), where the transverse and axial forces were correlated. This is likely because most or all of the transverse forces comes from the
impulse-like force transmission from the nasopharynx moving against the tip of the swab. As
the swab is not actively moving through the nasal cavity during this stage, Coloumb frictional forces would negligible The only variable
that seemed to have an impact on the mean value is the side of the nasal cavity the swab traversed, with greater forces occurring on the left side.

However, there is appears to be a difference in variance in axial force caused by the head motion level. The \textbf{Light} head motion has a
 less spread (stdv. 0.165 N) than \textbf{Medium} (0.27 N) or
 \textbf{Heavy} (0.365 N) motions. So as the head motion gets more erratic, so does the force transmission as the head moves into or away from
 the swab with increasing intensity.

\subsection{Stage 2: Measure vii) Oscillation coefficient}
\label{sec:s2_7}

\begin{figure}
      \includegraphics[width=\linewidth]{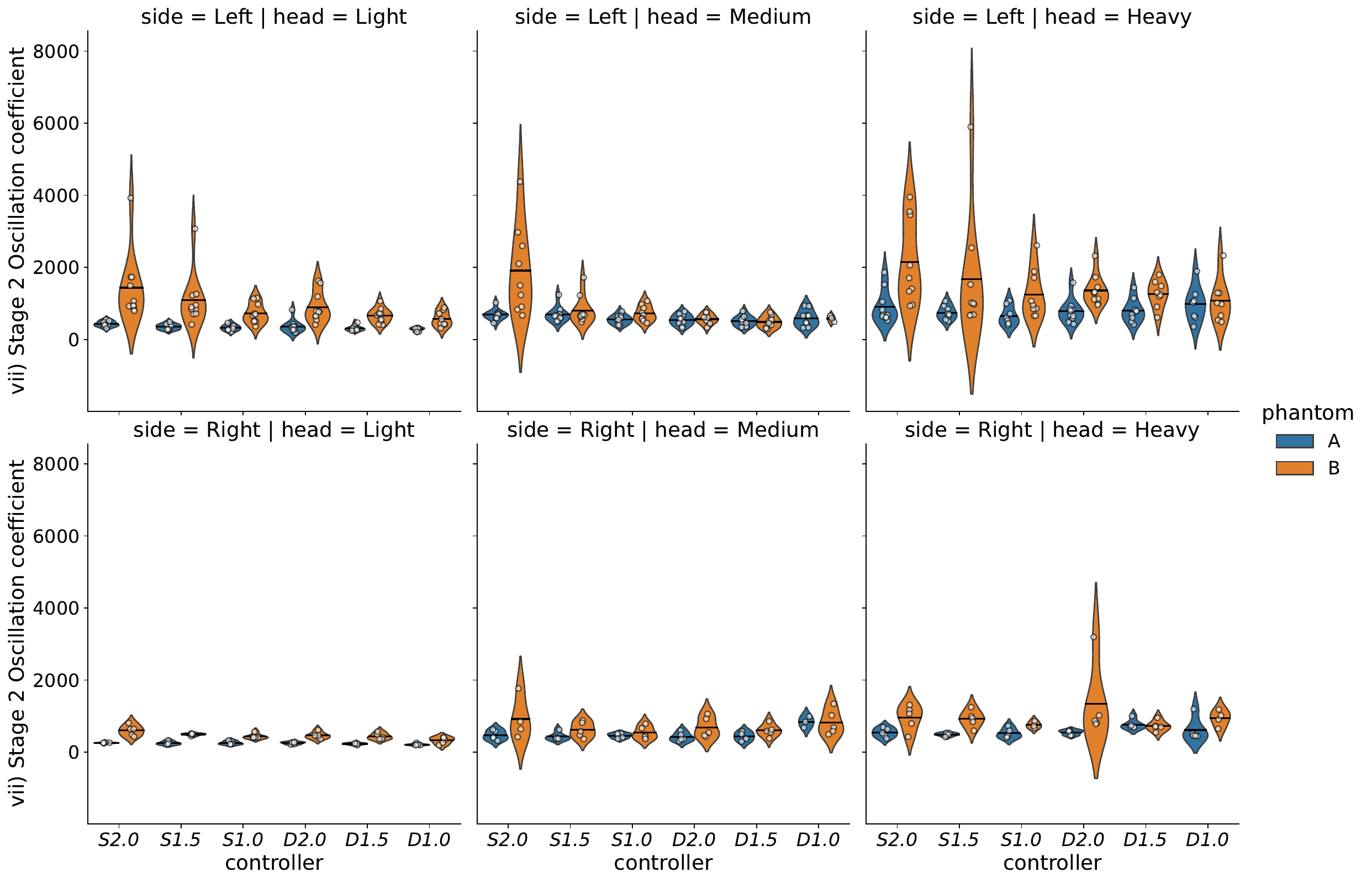}
      \caption[Measure vii) violin plot]{Measure vii) Violin plots showing distribution of oscillation coefficients during stage 2. See footnote \textsuperscript{\ref{ft:violin_expl}} in Section \ref{sec:metrics} for formatting.}
      \label{fig:s2_7}
    \end{figure}

As shown in Table \ref{tab:stage2_metrics} and Fig. \ref{fig:s2_7}, all four categorical variables have an effect on the oscillation
coefficient. First examining the 
controller variable, it appears that all the controller configurations had about the same oscillation,
with the glaring exception of \textit{S2.0}. In \textit{S2.0}, the controller with the highest gain at the
nasopharynx, there are instances where the oscillations dominated. For head motion, we see a steady
trend where heavy head motion leads to heavy oscillation. Finally, we see that both the left side of
the nostril and Phantom B tended to have higher oscillations than the right side and Phantom A
respectively.
Configurations with higher mean oscillation generally have higher variance as well, indicating that there is some \textit{on-off} attribute
towards oscillations getting triggered.

    \begin{table*}
      \caption{Statistics and summary of stage 2 measures.}
      \label{tab:stage2_metrics}
      \begin{center}
        \begin{tabular}{|c||c|c||c|c||c|c|}\hline
  & \multicolumn{6}{|c|}{\textbf{Four way ANOVA} }\\\hline
  & \multicolumn{2}{|c||}{\textbf{v) Transverse force}} & \multicolumn{2}{|c||}{\textbf{vi) Axial force}} & \multicolumn{2}{|c|}{\textbf{vii) Oscillation coef.}}\\\hline
  \underline{Variable} & \underline{F} & $\underline{p}$ & 
  \underline{F} & $\underline{p}$ &  \underline{F} & $\underline{p}$\\
  Control & 202  & $1.9 \times 10^{ -118 }$ * & 0.296  & $9.1 \times 10^{ -1 }$ 
  &  9.87  & $5.1 \times 10^{ -9 }$ *\\
  Head motion & 21.6 & $3.4 \times 10^{ -13 }$ * & 1.28 & $2.8 \times 10^{ -1 }$  & 28.7 & $3.6 \times 10^{ -17 }$ *\\
  Side & 20.4 & $7.8 \times 10^{ -6 }$ * & 21.1 & $5.5 \times 10^{ -6 }$ * &  38.6 & $1.1 \times 10^{ -9 }$ *\\
  Phantom & 24.9 & $8.1 \times 10^{ -7 }$ * & 0.645 & $4.2 \times 10^{ -1 }$  & 98.1 & $2.8 \times 10^{ -21 }$ * \\\hline
  \multicolumn{7}{|c|}{\textbf{Summary stats} } \\\hline
  & \multicolumn{2}{|c||}{\textbf{v) Transverse force} (N)} & \multicolumn{2}{|c||}{\textbf{vi) Axial force} (N)} & \multicolumn{2}{|c|}{\textbf{vii) Oscillation coef.}}\\\hline
  \underline{Control} & \underline{Mean} & \underline{Stdv.} &  \underline{Mean} & \underline{Stdv.} &  \underline{Mean} & \underline{Stdv.}\\
  \textit{S2.0} &   0.059   &   0.022 &    0.674   &   0.260 &    1028   &   895\\
  \textit{S1.5} &   0.069   &   0.019 &    0.693   &   0.249 &    768   &   707\\
  \textit{S1.0} &    0.098   &   0.024 &     0.688   &   0.284 &     617   &   367\\
  \textit{D2.0} &   0.109    &  0.030 &    0.687    &  0.260 &    689    &  470\\
  \textit{D1.5} &   0.139    &  0.037 &    0.660    &  0.261 &    604    &  326\\
  \textit{D1.0} &   0.186   &   0.055 &   0.699   &   0.329 &    643   &   404\\\hline
  \underline{Head motion} & \underline{Mean} & \underline{Stdv.} & 
  \underline{Mean} & \underline{Stdv.} &  \underline{Mean} & \underline{Stdv.}\\
  \textbf{None} & 0.100  &     0.040 & 0.639  &     0.096 &  555  &     669\\
  \textbf{Light} & 0.094  &    0.045  &  0.661  &    0.165  & 541  &    452 \\
  \textbf{Medium} & 0.111 & 0.051  &  0.716 & 0.270  & 674 & 461 \\
  \textbf{Heavy} &  0.120  &  0.060 &  0.679  &  0.365 &  1006  &  712 \\\hline
  \underline{Side}& \underline{Mean} & \underline{Stdv.} & \
  \underline{Mean} & \underline{Stdv.} &  \underline{Mean} & \underline{Stdv.}\\
  Left & 0.103  & 0.047 &  0.723  & 0.276 &  825  & 673\\
  Right & 0.115  & 0.061 & 0.611  & 0.250 &  554  & 332\\\hline
  \underline{Phantom} & \underline{Mean} & \underline{Stdv.} & 
  \underline{Mean} & \underline{Stdv.} &  \underline{Mean} & \underline{Stdv.}\\
  A &  0.101 &   0.049  &  0.693 &   0.264  &   523 &   276 \\
  B &  0.114 &   0.056  &  0.673 &   0.281  &  939 &   734 \\\hline
\end{tabular}

      \end{center}
    \end{table*}

\section{Discussion}
\label{sec:int:discussion}

We have done extensive testing of our proposed system to investigate the robustness of the
proposed system as well as to examine the tradeoffs of the different controller configurations
described in Section \ref{sec:compliant_controller}. Overall, we can see that higher
transverse gain will lead to lower transverse forces in stages 1 and 2 and lower axial forces in
stage 1. Having a high enough transverse gain is also important for being able to reach the
nasopharynx without getting stuck or prematurely triggering stage 2 from high axial
forces (Section \ref{sec:res_iv}). However, we observed that there is a risk for instability if the gain is
raised too high. Configuration \textit{S2.0}, the controller with the highest gain the nasopharynx, had
significant oscillations during stage 2 in many different trials. Therefore, the experimental
results indicate that it is beneficial to have a high gain at the beginning of the insertion, so the
controller can adjust to misalignment, motion, and avoid excessive
forces that would slow the trajectory progression down. But once stage 2 begins, the gain should be reduced to
avoid oscillations. This justifies the inspiration of our dynamic gain controllers, which can have
high gain at the beginning and taper down once it reaches the nasopharynx. Dynamic controller \textit{D2.0}
had a good trade-off in terms of all the measures, along with the static configuration \textit{S1.5}, which had
about equivalent performance.

In terms of the head motion simulation, there is a clear trend that heavier head motion causes greater disturbances to
the control system, as one would expect. With heavier motion we saw increased transverse (stage 1 and 2) and axial forces (stage 1),
increased durations for stage 1, and higher oscillation coefficients. For less well tuned controllers, we also saw this
had the impact of raising oscillation. Comparing the three configurations head motion configurations, \textbf{Heavy} moved quite rapidly and strayed far from the resting
point. The motion is extreme enough that it is unlikely a cooperating patient would move in such a manner.
The fact that the better tuned controllers were still able to handle these conditions builds confidence for its applications towards human subjects.
The \textbf{Light} motion would probably be the
 more realistic scenario for a cooperating patient, with \textbf{Medium} motion being the upper bound movement we would expect.

In our other work \cite{Lee2024-437}, we proposed the visual servo system for anthropometric faces and
conducted human trials to evaluate the system. There, we observed that the final angle could vary
with different faces due to the biases of the underlying CNN (3DDFA\_V2). This raises the question, could our system be able to compensate for this variability?
For a stationary phantom, the repeatability for our replacement visual servo system presented some variance, although from visual inspection it appeared to be lower
than the experiments from our previous study~\cite{Lee2024-437}.
The head motion was a larger source of placement variance that was comparable to the human trials, as the head would often move during the transition between the pre-contact and
contact stages, resulting in larger alignment discrepancies after the turn stage of the trajectory. The force feedback was generally able to
compensate for this misalignment, which leads us to believe the system compensate for the alignment errors observed during the human
trials. Of course, the early abort system is present in cases where the alignment is too extreme to work.

In terms of the statistical hypothesis tests used, one may recall that ANOVA testing typically makes three assumptions \cite{Maxwell2017-416}: a) independent
observations, b) normally distributed outcomes, c) and homogeneity of variances. It is clear by examining the figures in Section
\ref{sec:int_experiments} that several measures (e.g., vii)) violate the third assumption.
In most of the measures, a raise in mean was usually accompanied by raise in variance as well.
While statistically speaking, this would make it more difficult to reject the null hypothesis of unequal group means,
in our scenario this conclusion is less relevant. We mainly care if factor causes a change in behaviour for the controller, which is evident
when a configuration causes more outliers and higher variance in a measure.

The abort observer from Section \ref{sec:safety} was a useful component during the trials.
While we would not expect the level of movement from a cooperating patient, during the \textbf{Heavy} head motion trials there were several unrecorded instances
where the swab missed the nostril because the head moved rapidly before the swab could enter the
nostril at the beginning of the trajectory. In these cases the swab would make contact with the ``tissue''
outside the nostril and release the swab within 0.2 seconds of collision, before any real damage could happen. The maximum force
component was also useful for preventing the swab from becoming crumpled due to unpredictable heavy
head motion during one of the trials as noted in Table \ref{tab:siv_breakdown}.

\subsection{Avenues for improvement}

The best tuned controllers \textit{D2.0} and \textit{S1.5} were effective when subjected to the most challenging insertion conditions.
However, there is still some aspects that could be improved. One potential improvement could be the incorporation of additional components to further attenuate oscillations.
This could involve creating an oscillation observer via a short window Fourier transform, that could be used to detect if any oscillations
are occurring around the 2 Hz range. When oscillations are triggered from head motions or other external factors, they could be attenuated by
throttling the gain to lower levels. Altering the hardware to improve the latency between the force measurements and the commands/readings
from the robot could also be beneficial. 
While the head simulator was a useful formulation for generating random head motion, the motions were mostly continuous. One type of
motion that would be useful to simulate is the effects of rapid, impulsive motions like coughing or sneezing, that are quite probable to
occur in patients with swabs entering their nose. It would be interesting to see how the current controller would react to these events, and
if any additional modifications would be needed to compensate for them.

The total duration of the pre-contact (including the sentry phase) and contact phases averaged just below 70 seconds, which is comparable to other swab robots~\cite{Wang2024-423}~\cite{Sun2023-420}~\cite{Chen2022-391}. A human healthcare worker could likely perform the test much faster, and there is certainly potential to tune the trajectories for higher throughput. However, some caution is needed to ensure that the quality of the samples are not compromised by speeding up the procedures. In addition, patients may be more likely to panic if the robot is moving rapidly.

As discussed in our other work \cite{Lee2024-437},
integrating visual states into the insertion stage could be beneficial for keeping the swab
aligned.
This would be especially helpful during the first few seconds of the contact phase, when the swab has barely entered the nostril and the
head could easily move away without providing force feedback. Therefore, having a visual error term early on for the insertion could help
this issue substantially. There could be some transient benefit for having visual feedback once the swab is inserted further into the nasal
cavity, but there are technical limitations to consider. Once the camera gets too close to the face, the RGB-D camera will be unable to obtain depth measurements. This means that such a setup would have to rely on perspective within the
colour images or consider adding an external eye-to-hand camera. While an eye-to-hand camera could have some potential for estimating the
state of the target~\cite{Flandin2000-377}, it would require a method of identifying the relative pose from the new perspective.

\section{Conclusion}
\label{sec:int:conclusion}

    We demonstrated the application of a proposed multi-stage controller to perform the NP swab test on patient simulator with nasal cavity phantom undergoing head motion. We formulated a stochastic head motion model based on an Ornstein-Uhlenbeck process to simulate patients with adjustable levels of head motion. Our integrated controller uses a visual servo control law to align the swab next to the nose, and then transitions to a      velocity control law that is tuned according to the mechanical
    properties of the swab.     We also added subsystems to detect when the swab reached the nasopharynx, and to abort the action if safety criteria is violated. Extensive experiments were conducted to study the performance of different controller configurations under variable levels of head motions and alternate phantoms.

The proposed system performs well and was able to execute the test without any major mishaps. Thanks to the abort sequencing, the
robot can quickly disengage from the patient if the swab does not enter the nostril, large forces are measured, or by manual abort. Some
additional minor components could be considered to watch for oscillations and to possibly integrate visual states
during the contact phase. But overall, the better tuned configurations appear promising for application in human trials, given their
performance during the experiments.

There are several avenues to extend this work in future research. The obvious course is evaluating the proposed system in human trials and measuring the sampling efficiency by assessing PCR cycle thresholds. If the system were to be evaluated in clinical settings, some modifications would be needed to ensure the robot can be properly sanitized and to make it intuitive to non-technical staff. Some patients could be wary of interacting with a robot, so techniques from the field of social robotics should be investigated to help mitigate these factors. Finally, adding mobility to the platform and adapting it to perform additional types of close contact health tasks, such as temperature testing, ultrasound scanning, or needle work, could further push the envelope to improving general healthcare with robotics.

\bibliographystyle{IEEEtran}
\bibliography{ref}

\begin{thebibliography}{10}
\providecommand{\url}[1]{#1}
\csname url@rmstyle\endcsname
\providecommand{\newblock}{\relax}
\providecommand{\bibinfo}[2]{#2}
\providecommand\BIBentrySTDinterwordspacing{\spaceskip=0pt\relax}
\providecommand\BIBentryALTinterwordstretchfactor{4}
\providecommand\BIBentryALTinterwordspacing{\spaceskip=\fontdimen2\font plus
\BIBentryALTinterwordstretchfactor\fontdimen3\font minus
  \fontdimen4\font\relax}
\providecommand\BIBforeignlanguage[2]{{%
\expandafter\ifx\csname l@#1\endcsname\relax
\typeout{** WARNING: IEEEtran.bst: No hyphenation pattern has been}%
\typeout{** loaded for the language `#1'. Using the pattern for}%
\typeout{** the default language instead.}%
\else
\language=\csname l@#1\endcsname
\fi
#2}}

\bibitem{Hiebert2021-278}
N.~M. Hiebert, B.~A. Chen, and L.~J. Sowerby, ``Variability in instructions for
  performance of nasopharyngeal swabs across {Canada} in the era of {COVID-19}
  -- what type of swab is actually being performed?'' \emph{Journal of
  Otolaryngology - Head {\&} Neck Surgery}, vol.~50, no.~1, p.~5, Jan 2021.

\bibitem{Leber2020-173}
A.~L. Leber, J.~G. Lisby, G.~Hansen, R.~F. Relich, U.~V. Schneider, P.~Granato,
  S.~Young, J.~Pareja, I.~Hannet, and Y.-W. Tang, ``Multicenter evaluation of
  the qiastat-dx respiratory panel for detection of viruses and bacteria in
  nasopharyngeal swab specimens,'' \emph{Journal of Clinical Microbiology},
  vol.~58, no.~5, pp. e00\,155--20, 2020.

\bibitem{Wang2024-423}
Z.~Wang, S.~Li, X.~He, R.~Chen, L.~Liu, S.~Li, and H.~Liu, ``The {LINGCAI-II}
  system: A sampling robotic system for autonomous oropharyngeal swab
  sampling,'' \emph{IEEE Transactions on Medical Robotics and Bionics}, vol.~6,
  no.~2, pp. 448--459, 2024.

\bibitem{Sun2023-420}
F.~Sun, J.~Ma, T.~Liu, H.~Liu, and B.~Fang, ``Autonomous oropharyngeal-swab
  robot system for {COVID-19} pandemic,'' \emph{IEEE Transactions on Automation
  Science and Engineering}, vol.~20, no.~4, pp. 2469--2478, 2023.

\bibitem{Zhang2021-215}
H.~Zhang, Q.~Wang, C.~Chi, Y.~Chen, Z.~Mu, Z.~Li, Y.~Lan, and A.~Zhang,
  ``Design and implementation of a novel, intrinsically safe rigid-flexible
  coupling manipulator for {COVID-19} oropharyngeal swab sampling,'' in
  \emph{2021 IEEE International Conference on Robotics and Automation (ICRA)},
  2021.

\bibitem{Chen2022-391}
Y.~Chen, Q.~Wang, C.~Chi, C.~Wang, Q.~Gao, H.~Zhang, Z.~Li, Z.~Mu, R.~Xu,
  Z.~Sun, and H.~Qian, ``A collaborative robot for {COVID-19} oropharyngeal
  swabbing,'' \emph{Robotics and Autonomous Systems}, vol. 148, p. 103917,
  2022.

\bibitem{Pinninti2020-398}
S.~Pinninti, C.~Trieu, S.~K. Pati, M.~Latting, J.~Cooper, M.~C. Seleme,
  S.~Boppana, N.~Arora, W.~J. Britt, and S.~B. Boppana, ``{Comparing
  Nasopharyngeal and Midturbinate Nasal Swab Testing for the Identification of
  Severe Acute Respiratory Syndrome Coronavirus 2},'' \emph{Clinical Infectious
  Diseases}, vol.~72, no.~7, pp. 1253--1255, 06 2020.

\bibitem{Chen2022-313}
W.~Chen, Z.~Chen, Y.~Lu, H.~Cao, J.~Zhou, M.~C.~F. Tong, and Y.-H. Liu,
  ``Easy-to-deploy combined nasal/throat swab robot with sampling dexterity and
  resistance to external interference,'' \emph{IEEE Robotics and Automation
  Letters}, vol.~7, no.~4, pp. 9699--9706, 2022.

\bibitem{Maeng2022-314}
C.-Y. Maeng, J.~Yoon, D.-Y. Kim, J.~Lee, and Y.-J. Kim, ``Development of an
  inherently safe nasopharyngeal swab sampling robot using a force restriction
  mechanism,'' \emph{IEEE Robotics and Automation Letters}, vol.~7, no.~4, pp.
  11\,150--11\,157, 2022.

\bibitem{Shim2024-421}
S.~Shim and J.~Seo, ``Robotic system for nasopharyngeal swab sampling based on
  remote center of motion mechanism,'' \emph{International Journal of Computer
  Assisted Radiology and Surgery}, vol.~19, no.~3, pp. 395--403, Mar 2024.

\bibitem{Zhang2023-355}
T.~Zhang, S.~Zheng, C.~Liu, Y.~Yang, Z.~Sun, and T.~lun Lam, ``A dual-arm
  nasopharyngeal swab manipulation robot for polymerization chain reaction
  sampling,'' in \emph{Advances in Mechanism and Machine Science}, vol. 148,
  2023.

\bibitem{Haddadin2024-390}
S.~Haddadin, D.~Wilhelm, D.~Wahrmann, F.~Tenebruso, H.~Sadeghian, A.~Naceri,
  and S.~Haddadin, ``Autonomous swab robot for naso- and oropharyngeal
  {COVID-19} screening,'' \emph{Scientific Reports}, vol.~14, no.~1, p. 142,
  Jan 2024.

\bibitem{LimHyungsun2014-395}
L.~J. H.~L. Hyungsun, ``A method for optimal depth of the nasopharyngeal
  temperature probe: the philtrum to tragus distance,'' \emph{Korean J
  Anesthesiol}, vol.~66, no.~3, pp. 195--198, 2014.

\bibitem{Lee2022-309}
P.~Q. Lee, J.~S. Zelek, and K.~Mombaur, ``Simulating and optimizing
  nasopharyngeal swab insertion paths for use in robotics,'' in \emph{2022 9th
  IEEE RAS/EMBS International Conference for Biomedical Robotics and
  Biomechatronics (BioRob)}, 2022.

\bibitem{Lee2024-437}
------, ``Robotic eye-in-hand visual servo axially aligning nasopharyngeal
  swabs with the nasal cavity,'' \emph{arXiv preprint arXiv:2408.12437}, 2024.

\bibitem{Lee2024-436}
------, ``Collaborative robot arm inserting nasopharyngeal swabs with
  admittance control,'' \emph{arXiv preprint arXiv:2408.11688}, 2024.

\bibitem{Zhu2016-121}
X.~Zhu, Z.~Lei, X.~Liu, H.~Shi, and S.~Z. Li, ``Face alignment across large
  poses: A {3D} solution,'' in \emph{Proceedings of the IEEE Conference on
  Computer Vision and Pattern Recognition (CVPR)}, June 2016.

\bibitem{Ren2017-399}
S.~Ren, K.~He, R.~Girshick, and J.~Sun, ``Faster {R-CNN}: Towards real-time
  object detection with region proposal networks,'' \emph{Institute of
  Electrical and Electronics Engineers (IEEE)}, Jun 2017.

\bibitem{He2015-402}
K.~He, X.~Zhang, S.~Ren, and J.~Sun, ``Deep residual learning for image
  recognition,'' \emph{CoRR}, vol. abs/1512.03385, 2015.

\bibitem{Lin2014-401}
T.~Lin, M.~Maire, S.~J. Belongie, L.~D. Bourdev, R.~B. Girshick, J.~Hays,
  P.~Perona, D.~Ramanan, P.~Doll{\'{a}}r, and C.~L. Zitnick, ``Microsoft
  {COCO:} common objects in context,'' \emph{CoRR}, vol. abs/1405.0312, 2014.

\bibitem{Chen2019-400}
K.~Chen, J.~Wang, J.~Pang, Y.~Cao, Y.~Xiong, X.~Li, S.~Sun, W.~Feng, Z.~Liu,
  J.~Xu, Z.~Zhang, D.~Cheng, C.~Zhu, T.~Cheng, Q.~Zhao, B.~Li, X.~Lu, R.~Zhu,
  Y.~Wu, J.~Dai, J.~Wang, J.~Shi, W.~Ouyang, C.~C. Loy, and D.~Lin,
  ``{MMDetection}: Open mmlab detection toolbox and benchmark,'' \emph{arXiv
  preprint arXiv:1906.07155}, 2019.

\bibitem{Sagonas2013-375}
C.~Sagonas, G.~Tzimiropoulos, S.~Zafeiriou, and M.~Pantic, ``300 faces
  in-the-wild challenge: The first facial landmark localization challenge,'' in
  \emph{2013 IEEE International Conference on Computer Vision Workshops}, 2013.

\bibitem{Brossard2020-191}
M.~Brossard, A.~Barrau, and S.~Bonnabel, ``A code for unscented kalman
  filtering on manifolds (ukf-m),'' in \emph{2020 IEEE International Conference
  on Robotics and Automation (ICRA)}, 2020.

\bibitem{Kermorgant2014-365}
O.~Kermorgant and F.~Chaumette, ``Dealing with constraints in sensor-based
  robot control,'' \emph{IEEE Transactions on Robotics}, vol.~30, no.~1, pp.
  244--257, 2014.

\bibitem{Smith1999-379}
S.~Smith, \emph{Digital signal processing: a practical guide for engineers and
  scientists}.\hskip 1em plus 0.5em minus 0.4em\relax California Technical
  Publishing, 1999.

\bibitem{Powell1994-240}
M.~J.~D. Powell, \emph{A Direct Search Optimization Method That Models the
  Objective and Constraint Functions by Linear Interpolation}.\hskip 1em plus
  0.5em minus 0.4em\relax Springer Netherlands, 1994, pp. 51--67.

\bibitem{Sananes2020-91}
N.~Sanan{\`e}s, M.~Lodi, A.~Koch, L.~Lecointre, A.~Sanan{\`e}s, N.~Lefebvre,
  and C.~Debry, ``{3D}-printed simulator for nasopharyngeal swab collection for
  {COVID-19},'' \emph{European Archives of Oto-Rhino-Laryngology}, Nov 2020.

\bibitem{AAstrom2008-272}
K.~J. {\AA}str{\"o}m and R.~M. Murray, \emph{Feedback systems: an introduction
  for scientists and engineers}.\hskip 1em plus 0.5em minus 0.4em\relax
  Princeton university press Princeton, NJ, 2008.

\bibitem{Bauchau2009-190}
O.~A. Bauchau and J.~I. Craig, \emph{Three-dimensional beam theory}.\hskip 1em
  plus 0.5em minus 0.4em\relax Springer Netherlands, 2009, pp. 223--259.

\bibitem{Liu2009-171}
Y.~Liu, M.~R. Johnson, E.~A. Matida, S.~Kherani, and J.~Marsan, ``Creation of a
  standardized geometry of the human nasal cavity,'' \emph{Journal of Applied
  Physiology}, vol. 106, no.~3, pp. 784--795, 2009.

\bibitem{Keustermans2018-172}
W.~Keustermans, T.~Huysmans, F.~Danckaers, A.~Zarowski, B.~Schmelzer,
  J.~Sijbers, and J.~J.~J. Dirckx, ``High quality statistical shape modelling
  of the human nasal cavity and applications,'' \emph{Royal Society Open
  Science}, vol.~5, no.~12, p. 181558, 2018.

\bibitem{Reznikov1995-174}
M.~Reznikov, T.~K. Blackmore, J.~J. Finlay-Jones, and D.~L. Gordon,
  ``Comparison of nasopharyngeal aspirates and throat swab specimens in a
  polymerase chain reaction-based test formycoplasma pneumoniae,''
  \emph{European Journal of Clinical Microbiology and Infectious Diseases},
  vol.~14, no.~1, pp. 58--61, Jan 1995.

\bibitem{thesis}
P.~Q. Lee, \emph{Autonomous Robotic System Conducting Nasopharyngeal
  Swabbing}.\hskip 1em plus 0.5em minus 0.4em\relax University of Waterloo,
  2024, ch. Hardware used and developed for nasopharyngeal swabbing, {PhD}
  dissertation.

\bibitem{Park2021-301}
C.~Park, I.~Choi, J.~Roh, S.~Y. Lim, S.-H. Kim, J.~Lee, and S.~Yang,
  ``Evaluation of applied force during nasopharyngeal swab sampling using
  handheld sensorized instrument,'' in \emph{2021 43rd Annual International
  Conference of the IEEE Engineering in Medicine \& Biology Society (EMBC)},
  2021.

\bibitem{Sadoughi2016-361}
N.~Sadoughi and C.~Busso, \emph{Head Motion Generation}.\hskip 1em plus 0.5em
  minus 0.4em\relax Springer International Publishing, 2016, pp. 1--25.

\bibitem{BenYoussef2013-362}
A.~B. Youssef, H.~Shimodaira, and D.~A. Braude, ``Head motion analysis and
  synthesis over different tasks,'' in \emph{Intelligent Virtual Agents}, 2013.

\bibitem{Mittal2023-383}
T.~Mittal, Z.~Aldeneh, M.~Fedzechkina, A.~Ranjan, and B.-J. Theobald,
  ``Naturalistic head motion generation from speech,'' in \emph{ICASSP 2023 -
  2023 IEEE International Conference on Acoustics, Speech and Signal Processing
  (ICASSP)}, 2023.

\bibitem{Chen2020-384}
L.~Chen, G.~Cui, C.~Liu, Z.~Li, Z.~Kou, Y.~Xu, and C.~Xu, ``Talking-head
  generation with rhythmic head motion,'' in \emph{Computer Vision -- ECCV
  2020}, 2020.

\bibitem{Agarwal2016-385}
P.~Agarwal, S.~A. Moubayed, A.~Alspach, J.~Kim, E.~J. Carter, J.~F. Lehman, and
  K.~Yamane, ``Imitating human movement with teleoperated robotic head,'' in
  \emph{2016 25th IEEE International Symposium on Robot and Human Interactive
  Communication (RO-MAN)}, 2016.

\bibitem{Mulavara2002-387}
A.~P. Mulavara, M.~C. Verstraete, and J.~J. Bloomberg, ``Modulation of head
  movement control in humans during treadmill walking,'' \emph{Gait \&
  Posture}, vol.~16, no.~3, pp. 271--282, 2002.

\bibitem{Antonelli2014-388}
M.~Antonelli, A.~P. del Pobil, and M.~Rucci, ``Bayesian multimodal integration
  in a robot replicating human head and eye movements,'' in \emph{2014 IEEE
  International Conference on Robotics and Automation (ICRA)}, 2014.

\bibitem{Hirasaki1999-386}
E.~Hirasaki, S.~T. Moore, T.~Raphan, and B.~Cohen, ``Effects of walking
  velocity on vertical head and body movements during locomotion,''
  \emph{Experimental Brain Research}, vol. 127, no.~2, pp. 117--130, Jul 1999.

\bibitem{Leva1996-350}
P.~{de Leva}, ``Adjustments to zatsiorsky-seluyanov's segment inertia
  parameters,'' \emph{Journal of Biomechanics}, vol.~29, no.~9, pp. 1223--1230,
  1996.

\bibitem{Heuring1999-363}
J.~Heuring and D.~Murray, ``Modeling and copying human head movements,''
  \emph{IEEE Transactions on Robotics and Automation}, vol.~15, no.~6, pp.
  1095--1108, 1999.

\bibitem{Uhlenbeck1930-364}
G.~E. Uhlenbeck and L.~S. Ornstein, ``On the theory of the brownian motion,''
  \emph{American Physical Society}, vol.~36, pp. 823--841, Sep 1930.

\bibitem{Kuo2020-357}
C.-F.~J. Kuo, Y.-S. Leu, D.-J. Hu, C.-C. Huang, J.-J. Siao, and K.~B.~P. Leon,
  ``Application of intelligent automatic segmentation and 3d reconstruction of
  inferior turbinate and maxillary sinus from computed tomography and analyze
  the relationship between volume and nasal lesion,'' \emph{Biomedical Signal
  Processing and Control}, vol.~57, p. 101660, 2020.

\bibitem{Smith2018-359}
D.~H. Smith, C.~D. Brook, S.~Virani, and M.~P. Platt, ``The inferior turbinate:
  An autonomic organ,'' \emph{American Journal of Otolaryngology}, vol.~39,
  no.~6, pp. 771--775, 2018.

\bibitem{Doss2023-433}
A.~S. {Arockia Doss}, P.~K. Lingampally, G.~M.~T. Nguyen, and D.~Schilberg, ``A
  comprehensive review of wearable assistive robotic devices used for head and
  neck rehabilitation,'' \emph{Results in Engineering}, vol.~19, p. 101306,
  2023.

\bibitem{Ferrario2002-434}
V.~F. Ferrario, C.~Sforza, G.~Serrao, G.~Grassi, and E.~Mossi, ``Active range
  of motion of the head and cervical spine: a three-dimensional investigation
  in healthy young adults,'' \emph{Journal of Orthopaedic Research}, vol.~20,
  no.~1, pp. 122--129, 2002.

\bibitem{Dunn1961-382}
O.~J. Dunn, ``Multiple comparisons among means,'' \emph{[American Statistical
  Association, Taylor {\&} Francis, Ltd.]}, vol.~56, no. 293, pp. 52--64,
  2024/05/16/ 1961.

\bibitem{Maxwell2017-416}
S.~E. Maxwell, H.~D. Delaney, and K.~Kelley, \emph{Designing experiments and
  analyzing data: A model comparison perspective}.\hskip 1em plus 0.5em minus
  0.4em\relax Routledge, 2017.

\bibitem{Flandin2000-377}
G.~Flandin, F.~Chaumette, and E.~Marchand, ``{Eye-in-hand / eye-to-hand
  cooperation for visual servoing},'' in \emph{{IEEE Int. Conf. on Robotics and
  Automation, ICRA'00}}, vol.~3, 2000.

\end{thebibliography}

\end{document}